\documentclass[journal]{IEEEtran}
\usepackage{graphicx}
\usepackage{epsfig}
\usepackage{color}
\usepackage{soul}
\usepackage{float}
\usepackage{booktabs}
\usepackage{mathrsfs}
\usepackage{algorithm}
\usepackage{algorithmic}
\usepackage{epstopdf}
\usepackage{amsmath,amssymb} 
\newtheorem{prop}{Proposition}
\usepackage{xcolor}
\usepackage{subfigure}

\ifCLASSINFOpdf
\else
\fi
\begin{document}
%
\title{Robust and Discriminative Labeling for Multi-label Active Learning Based on Maximum Correntropy Criterion }
%
%
%

\author{Bo~Du,~
        Zengmao~Wang,~
        Lefei~Zhang,~
        Liangpei~Zhang,~
        Dacheng~Tao,
\thanks{This work was supported in part by the National Natural Science Foundation of China under Grants 61471274, 41431175 and 61401317, by the Natural Science Foundation of Hubei Province under Grant 2014CFB193, by the Fundamental Research Funds for the Central Universities under Grant 2042014kf0239, and by Australian Research Council Projects FT-130101457, DP-140102164, LP-150100671.
}
\thanks{B. Du, Z. Wang and L. Zhang are with the School of Computer, Wuhan University, Wuhan
430079, China (email:gunspace@163.com; wzm902009@gmail.com; zhanglefei@whu.edu.cn).(Corresponding Author: Lefei Zhang)}
\thanks{L. Zhang is with the State Key Laboratory of Information Engineering
in Surveying, Mapping, and Remote Sensing, Wuhan University, Wuhan
430072, China (email:zlp62@whu.edu.cn).}
\thanks{D. Tao is with the School of Information Technologies and the Faculty of Engineering and Information Technologies, University of Sydney, J12/318 Cleveland St, Darlington NSW 2008, Australia (email: dacheng.tao@sydney.edu.au).}
\thanks{\copyright 20XX IEEE. Personal use of this material is permitted. Permission from IEEE must be obtained for all other uses, in any current or future media, including reprinting/republishing this material for advertising or promotional purposes, creating new collective works, for resale or redistribution to servers or lists, or reuse of any copyrighted component of this work in other works.}
}

\maketitle

\begin{abstract}
Multi-label learning draws great interests in many real world applications. It is a highly costly task to assign many labels by the oracle for one instance. Meanwhile, it is also hard to build a good model without diagnosing discriminative labels. Can we reduce the label costs and improve the ability to train a good model for multi-label learning simultaneously?

Active learning addresses the less training samples problem by querying the most valuable samples to achieve a better performance with little costs. In multi-label active learning, some researches have been done for querying the relevant labels with less training samples or querying all labels without diagnosing the discriminative information. They all cannot effectively handle the outlier labels for the measurement of uncertainty. Since Maximum Correntropy Criterion (MCC) provides a robust analysis for outliers in many machine learning and data mining algorithms, in this paper, we derive a robust multi-label active learning algorithm based on MCC by merging uncertainty and representativeness, and propose an efficient alternating optimization method to solve it. With MCC, our method can eliminate the influence of outlier labels that are not discriminative to measure the uncertainty. To make further improvement on the ability of information measurement, we merge uncertainty and representativeness with the prediction labels of unknown data. It can not only enhance the uncertainty but also improve the similarity measurement of multi-label data with labels information. Experiments on benchmark multi-label data sets have shown a superior performance than the state-of-the-art methods.

\end{abstract}


%
\IEEEpeerreviewmaketitle

\section{Introduction}
%
%
%
%
\IEEEPARstart {M}{achine} learning is the topic of the day. However, less training samples problem is always the challenge problem in machine learning fields\cite{s1}. Especially nowadays amount of data is generated quickly within a short period time. Active learning as a subfield of machine learning is an effective machine learning approach to address the less training samples problem in classification. It has been elaborately developed for various classification tasks by querying the most valuable samples, which is an iterative loop to find the most valuable samples for the 'oracle' to label, and gradually improves the models' generalization ability until the convergence condition is satisfied \cite{s1}. In general, there are two motivations behind the design of a practical active learning algorithm, namely, uncertainty and representativeness\cite{s2,s3,s4}. Uncertainty is the criterion used to select the samples that can help to improve the generalization ability of classification models, ensuring that the classification results of unknown data are more reliable. Representativeness measures the overall patterns of the unlabeled data to prevent the bias of a classification model with few or no initial labeled data. No matter which kind of active learning method is used, the key lies on how to select the most valuable samples, which is referred as the query function.

Among all the tasks, such as object recognition, scene classification\cite{s11}, and image retrieval, multi-label classification, which aims to assign each instance with multiple labels, may be the most difficult one\cite{s5,s6,s7}.
For each training sample, different combinations of labels need to be considered. Compared with single-label classification,
the labeling of multi-label classification is more costly\cite{s25}. Currently, multi-label learning has been successfully applied in machine learning and computer vision fields,
including web classification\cite{s12}, video annotation\cite{s13}, and so on\cite{s8,s9,s10}.
To solve the classification tasks, many types of machine learning algorithms have been developed. However, less training samples is not solved in many of these techniques, and they all face such a problem. Hence,
active learning has become even more important to solve the less training samples problem, reducing the costs of the various classification tasks.
\begin{figure*}[ht]\label{fig1}
\begin{center}
\epsfig{file = 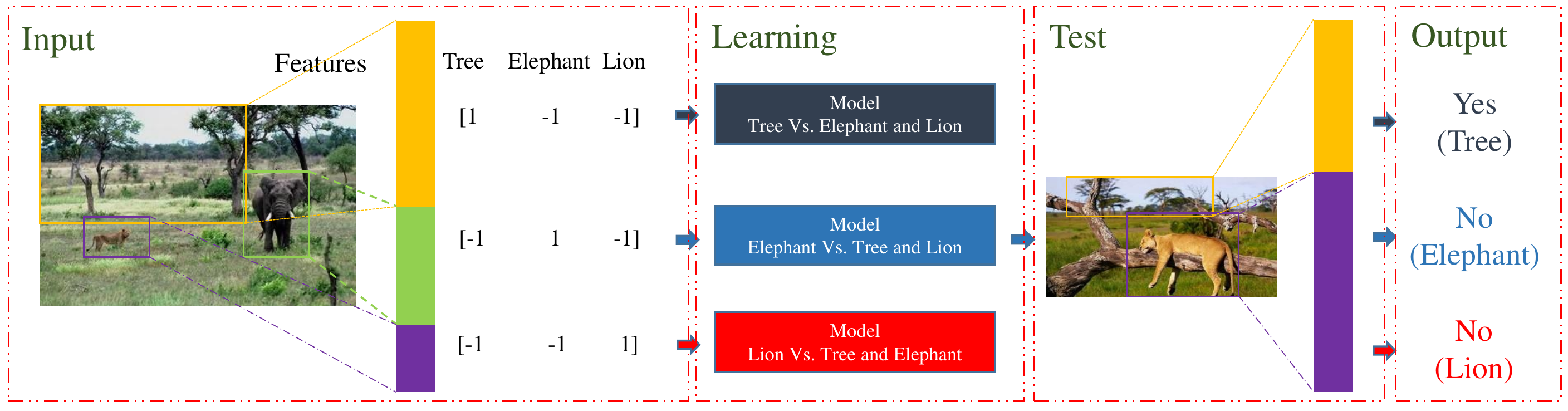, height = 1.8 in, keepaspectratio}
\caption{The influence of outlier label in the learning process.}
\end{center}
\end{figure*}

State-of-the-art multi-label active learning algorithms can be classified into three categories based on the query function. One category relies on the labeled data to design a query function with uncertainty\cite{s14,s15,s16}. For these methods, the design of query function ignores the structural information in the large-scale unlabeled data, leading to a serious sample bias or an undesirable performance. To eliminate this problem, the second category, which depends on the representativeness, has been developed\cite{s17,s18,s19}. In these approaches, the structural information of the unlabeled data is elaborately considered, but the discriminative (uncertain) information is discarded. Therefore, a large number of samples would be required before an optimal boundary is found. Since utilizing either the uncertainty criterion or the representativeness criterion may not achieve a desirable performance, the third category which combines both criteria\cite{s3,s4,s20} born naturally. These methods are either heuristic in designing the specific query criterion or ad hoc in measuring the uncertainty and representativeness of the samples. Although it is effective, the two parts are independent. Hence, the uncertainty still just relies on the limited labeled samples, and the information of two criteria are not enhanced. Most importantly, they ignored the outlier labels that exist in multi-label classification when designed a query function for active learning.

 However, the outlier labels have significant influence on the measurement of uncertainty and representativeness in multi-label learning. In the following, we will discuss the outlier label and its negative influence on the measurement of uncertainty and representativeness in detail.

Fig. 1 shows a simple example about the influence of outlier labels. As the input, we annotate the image with three labels, namely tree, elephant and lion. Hence, the feature of image is combined with three parts, the feature of tree, the feature of elephant and the feature of lion. Intuitively, in the image feature, the feature of tree is much more than elephant and lion, and the feature of lion is the least. If we use the image with the three labels to learn a lion/ non-lion binary classification model, the model would actually depend on the tree¡¯s and elephant¡¯s features rather than the lion¡¯s. Thus it would be a biased model for classifying the lion and the non-lions. Given the test image where a lion covers the most regions in the image, the trained model would not recognize the lion. If we use such a model to measure the uncertainty in active learning, it may cause wrong measurement for the images with lion label. We name the lion label in the input image as an outlier label.

Furthermore, we present the formal definition of an outlier label. Denote $(x,y_1,y_2 )$ as the sample-label pairs of an instance. $y_2$ is the most relevant label of the instance $x$, and $y_2$ is much more relevant to $x$ than $y_1$. Define $y_1$ as the outlier label, if it has the two properties. The first one is that $y_1$ is a relevant label to the instance $x$, and the second one is that compared with the most relevant label $y_2$, $y_1$ is much less relevant to $x$ than $y_2$. Fig. 2 shows the two properties, and we can understand the outlier label easier from it. According to the definition of outlier label, the lion is naturally treated as the outlier label. Since the feature of lion is not very obvious, if we treat the lion as a positive label, and use the image in Fig. 2 to learn a model, the model would not be able to effectively query an informative sample. Hence, if the influence of the outlier label can be avoided or decreased when we query the most informative sample, it would be very useful to build a promising model for classification. The definition of the outlier label also fits the fact that the outlier label may not be paid attention by the oracle at the first glance. In Fig. 2, the tree can be recognized at the first glance by the oracle, but the lion is very veiled and may be ignored with careless. The definition of outlier label is also consistent with the query types proposed in \cite{s21}.

\begin{figure}\label{fig2}
\begin{center}
\epsfig{file = 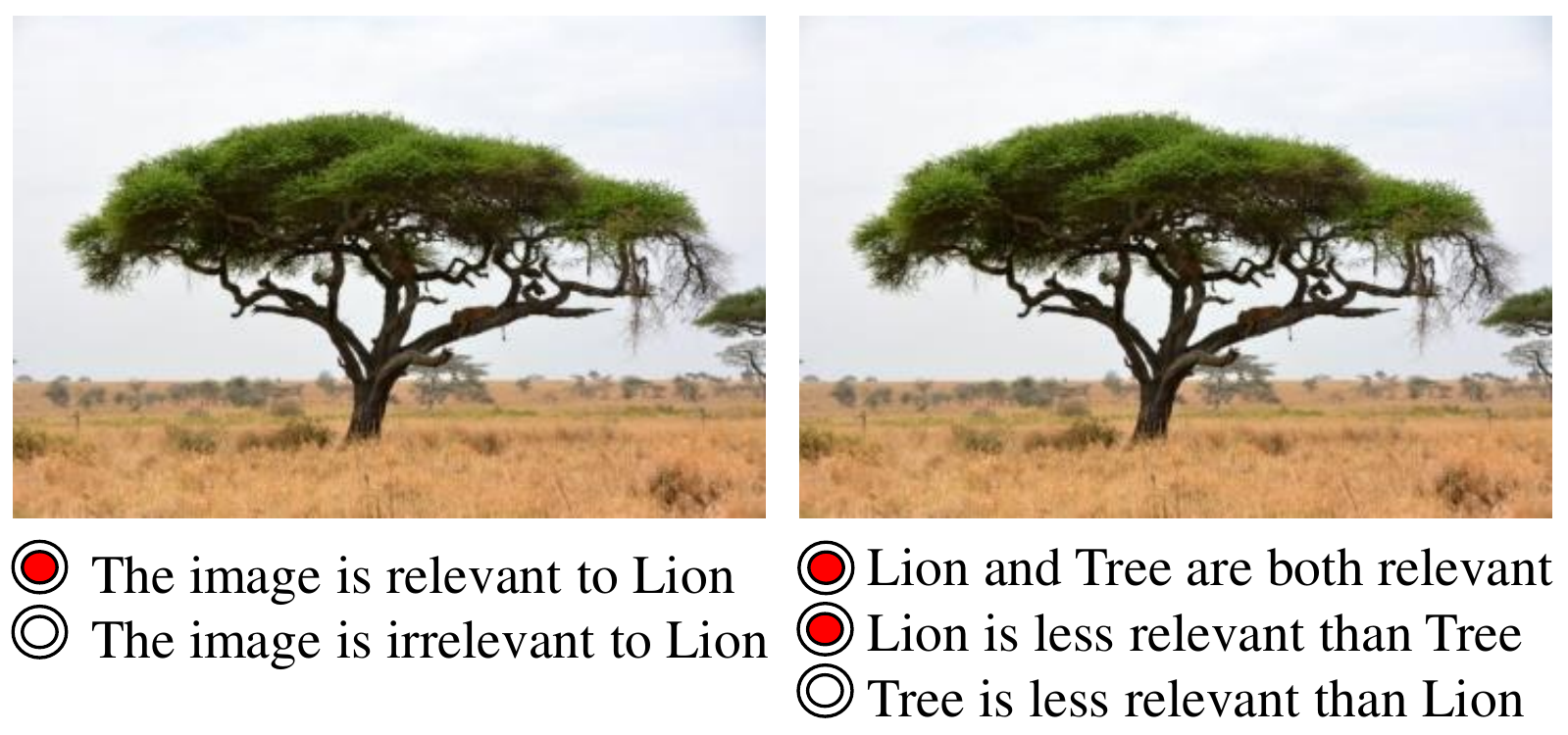, height = 1.5 in, keepaspectratio}
\caption{The interface of two properties for outlier labels. Left: The outlier label (Lion) is relevant to the image; right: the outlier (Lion) is much less relevant to the image than the most relevant label (Tree) is.}
\end{center}
\end{figure}
For two multi-label images, if they have the same labels with different outlier labels, this may lead to the result that the features of the two images have a large difference. Therefore, it is very hard to diagnose the similarity based on features between two instances with different outlier labels. In Fig. 3, we provide a simple example to show such a problem. We present the similarity between the sift features with Gaussian kernel, and the labels similarity based on MCC. In Fig. 3, the similarity between image 1 and image 2 should be larger than the similarity between image 2 and image 3, since the labels in image 1 and image 2 are exactly the same. However, the result is opposite when the similarity is measured with their sift features. The outlier label is lion in image 1, and tree trunk in image 2. The two outlier labels will largely increase the features' difference of the two images. In summary, the measurement of uncertainty and representativeness would be deteriorated with the outlier labels.
\begin{figure*}\label{fig3}
\begin{center}
\epsfig{file = 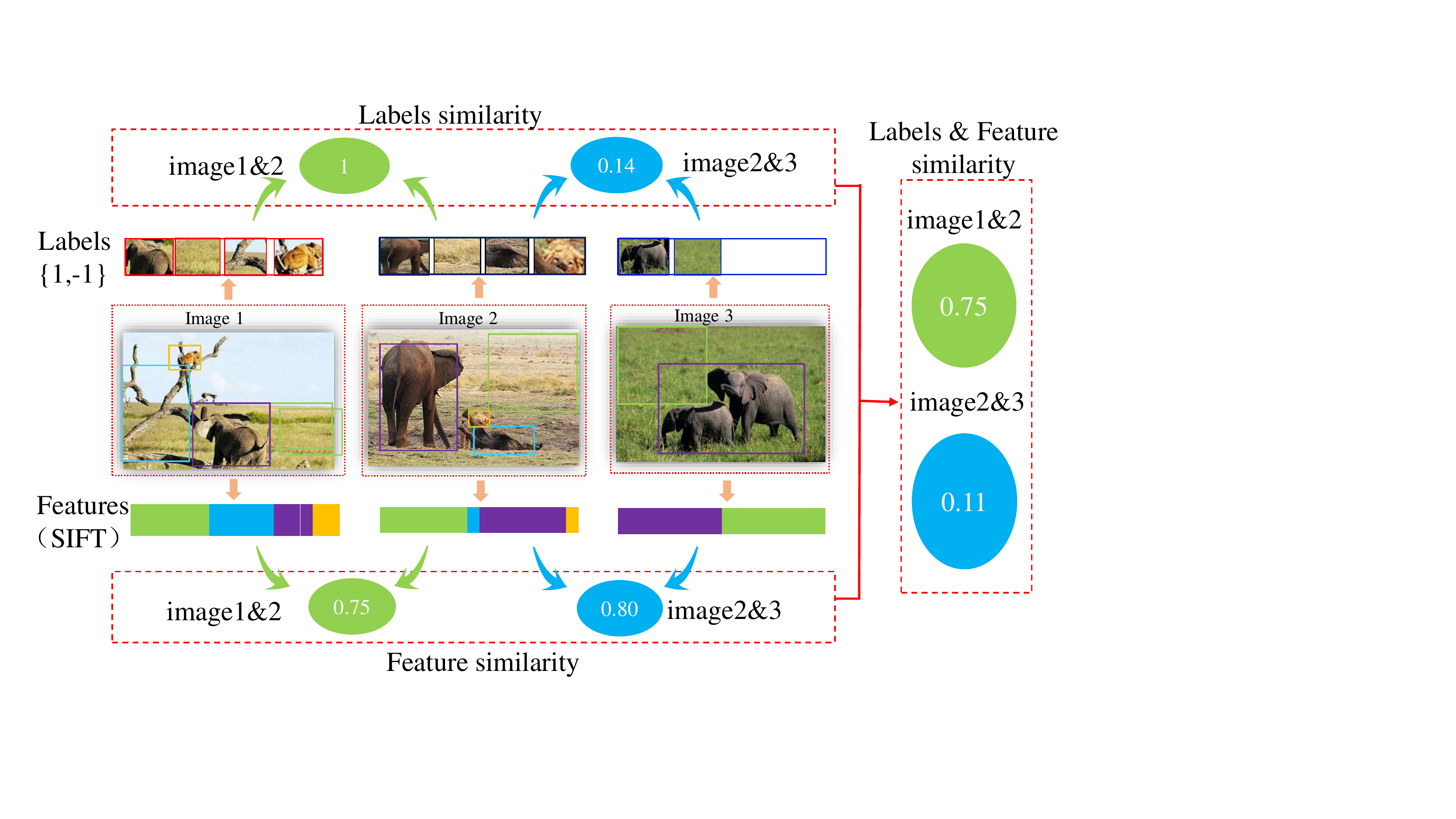,  height = 2.5 in, keepaspectratio}
\caption{The influence of the outlier labels for the measurement of similarity}
\end{center}
\end{figure*}

To address the above problems, in this paper, we proposed a robust multi-label active learning (RMLAL) algorithm, which effectively merges the uncertainty and the representativeness based on MCC.

As to robustness, the correntropy has been proved promising in information theoretic learning (ITL)\cite{s22} and can efficiently handle large outliers\cite{s23,s24}. In conventional active learning algorithms, the mean square error (MSE) cannot easily control the large errors caused by the outliers. For example, in Fig. 2, there are two labels for the image: tree and lion. If we use the lion model to learn the image, the prediction value must be very far from the lion's label. If we use the MSE loss to measure the loss between the prediction value and the label, a large error may be introduced, since MSE extends the error with square. MCC calculates the loss between the prediction value and the label with a kernel function. If a large enough error is introduced, the value of MCC is almost equal to zero. Hence, the influence of the large error will be restrained. We therefore replace MSE loss with MCC in the minimum margin model in the proposed formulation. In this way, the proposed method is able to eliminate the influence of the outlier labels, making the query function more robust.

As to discriminative labeling, we use the MCC to measure the loss between the true label and the prediction label. MCC can improve the most discriminative information and suppress the useless information or unexpected information. Hence, with MCC in the proposed method, if the label is not an outlier label, it will play an important role in the query model construction. Otherwise, the model will decrease the influence of the outlier label to measure the uncertainty. With such an approach, the discriminative labels' effects are improved and the outlier labels' are suppressed. Thus, the discriminative labeling can be achieved.

 As to representativeness, we mix the prediction labels of the unlabeled data with the MCC as the representativeness. As is shown in Fig. 3, although the samples have the same labels, their outlier labels are different, making their features distinguishing. If we just use the corresponding features to measure the similarity, it will lead to a wrong diagnosis. Hence, we propose to use the combination of labels' and features' similarity to define the consistency. The combination makes the measurement of representativeness more general. To decrease the computational complexity of the proposed method, the half-quadratic optimization technique is adopted to optimize the MCC.

  The contributions of our work can be summarized as follows:
\begin{itemize}
\item	To the best of our knowledge, it is the first work to focus on the outlier labels in multi-label active learning. We find a robust and effective query model for multi-label active learning.
\item	The prediction labels of the unlabeled data and the labels of the labeled data are utilized with MCC to merge the uncertain and representative information, deriving an approach to make the uncertain information more precise.
\item	The proposed representative measurement considers labels similarity by MCC. It can handle the outlier labels effectively and make the similarity more accurate for multi-label data. Meanwhile, a new way is provided to merge representativeness into uncertainty.
\end{itemize}

The rest of the paper is organized as follows: Section 2 briefly introduces the related works. Section 3 defines and discusses a new objective function for robust multi-label active learning, and then proposes an algorithm based on half-quadratic optimization. Section 4 evaluates our method on several benchmark multi-label data sets. Finally, we summarize the paper in Section 5.
\section{Related works}
Multi-label problem is universal in the real world, so that multi-label classification has drawn great interests in many fields. For a multi-label instance, it needs the human annotator to consider all the relevant labels. Hence, the labeling of multi-label tasks is more costly than single label learning, but the research of active learning on multi-label learning is still less.

 In multi-label learning, one instance is corresponding to more than one labels. To solve a multi-label problem, it is a direct way to convert the multi-label problem into several binary problems\cite{s39,s40}. In these approaches, the uncertainty is measured for each label, and then a combined strategy is adopted to measure the uncertainty of the instance. \cite{s39} trained a binary SVM model for each label and combined them with different strategies for the instance selection. \cite{s40} predicted the number of relevant labels for each instance by a logistic regression, and then adopted the SVM models to minimize the expected loss for the instance selection. Recently, \cite{s8} adopted the mutual information to design the selection criterion for Bayesian multi-label active learning, and \cite{s26} selected the valuable instances by minimizing the empirical risk. Other works have been done by combining the informativeness and representativeness together for a better query\cite{s27,s38}. \cite{s27} combined the label cardinality inconsistency and the separation margin with a tradeoff parameter. \cite{s38} took into account the cluster structure of the unlabeled instances as well as the class assignments of the labeled examples for a better selection of instances. All the above algorithms were designed to query all the labels of the selected instances without diagnosing discriminative labels. Another kind of approaches were developed to query the label-instance pairs with relevant label and instance at each iteration\cite{s21,s28,s29}. \cite{s21} queried the most relevant label based on the types. \cite{s29} selected label-instance pairs based on a label ranking model. In these approaches, the most relevant label is assigned to the instance, and some relevant labels may be lost with the limited query labels. Therefore, it may need to query much more label-instance pairs to achieve a good performance. It is proved that considering the combination of informativeness and representativeness is very effective in active learning. We adopt this strategy in the paper.

No matter whether selecting the instance by all the labels or by the label-instance pairs, most of the active learning algorithms only selected the uncertain instance based on very limited samples, and ignored the labels information. For example, given all the labels to one instance, if the instance has many outlier labels, such instance may decrease the performance of the classification task. To address these problems, we use the prediction labels of the unlabeled data to enhance the uncertain measurement and adopt the MCC to consider the relevant labels as many as possible except the outlier labels. As far to our knowledge, it is the first time to adopt the MCC in multi-label active learning with data labels for query.
\section{Multi-label active learning}
Suppose we are given a multi-label data set $D = \{x_i, x_2,...,x_n\}$ with $n$ samples and $C$ possible labels for each sample. Initially, we label $l$ samples in $D$. Without loss of generality, we denote the $l$ labeled samples as set $L = \{(x_1,y_1), (x_2,y_2),..., (x_l,y_l)\}$, where $y_i = (y_{i1},y_{i2},..., y_{iC})$ is the labels set for sample $x_i$, with $y_{ik}\in \{-1,1\}$; and the remaining $u = n - l$ unlabeled samples are denoted as set $U = \{x_{l+1}, x_{l+2},..., x_{l+u}\}$. It is the candidate set for active learning. Moreover, we denote $x_q$ as the sample that we want to query in the active learning process, and define that $y_L = \{y_1, y_2,...,y_l\}$ is the labels matrix for the labeled data. In the following discussion, the symbols are used as above.
\subsection{Maximum Correntropy Criterion}
In multi-label classification tasks, the outlier labels are the great challenge to train a precise classifier, mainly due to the unpredictable nature of the errors (bias) caused by those outliers. In active learning, in particular, the limited labeled samples with outliers easily lead to a great bias. This would directly lead to the bias of uncertain information, furthermore make the query instances undesirable or even lead to bad performances. Recently, the concept of correntropy was firstly proposed in information theoretic learning (ITL) and it had drawn much attention in the signal processing and machine learning community for robust analysis, which can effectively handle the outliers\cite{s30,s31}. In fact, correntropy is a similarity measurement between two arbitrary random variables $a$ and $b$\cite{s23,s30}, defined by
\begin{equation}\label{formula1}
{V_\sigma }\left( {a,b} \right) = E\left[ {{K_\sigma }\left( {a,b} \right)} \right]
\end{equation}
where $K_\sigma(\cdot)$ is the kernel function that satisfies Mercer theory and $E[\cdot]$ is the expectation operator. We can observe that the definition of correntropy is based on the kernel method. Hence, it has the same advantages that the kernel technique owns. However, different from the conventional kernel based methods, correntropy works independently with pairwise samples and has a strong theoretical foundation\cite{s23}. With such a definition, the properties of correntropy are symmetric, positive and bounded.

However, in the real world applications, the joint probability density function of $a$ and $b$ is unknown, and the available data in $a$ and $b$ are finite. We define the finite number of available data in $a$ and $b$ is $m$, and the data set is denoted as $\{a_i,b_i\}_{i=1}^m$. Thus, the sample estimator of correntropy is usually adopted by
\begin{equation}\label{formula2}
{\hat V_{m,\sigma} }\left( {a,b} \right) = \frac{1}{m}\sum\limits_{i = 1}^n {{K_\sigma }\left( {{a_i},{b_i}} \right)}
\end{equation}
where $K_\sigma(a_i,b_i)$ is Gaussian kernel function $g(a_i,b_i) \triangleq exp(-||a_i-b_i||^2)/2\sigma^2)$. According to\cite{s23,s30}, the correntropy between $a$ and $b$ is given by
\begin{equation}\label{formula3}
 \max \limits_{p^{'}} \frac{1}{m}\sum\limits_{i = 1}^m {g\left( {a_i,b_i} \right)}
\end{equation}

The objective function (3) is called MCC, where $p^{'}$ is the auxiliary parameter to be specified in Proposition \ref{Proposition2}. Compared with MSE, which is a global metric, the correntropy is a local metric. That means the correntropy value is mainly determined by the kernel function along the line $a$ = $b$.
\subsection{Multi-label Active learning based on MCC}
Usually, in active learning methods, the uncertainty is measured according to the labeled data whereas the representativeness according to the unlabeled data. In this paper, we propose a novel approach to merge the uncertainty and representativeness of instances in active learning based on MCC. Mathematically, it is formulated as an optimization
problem \emph{w.r.t.} the classifier $f$ and the query sample $x_q$:
\begin{equation}\label{formula28}
\begin{split}
\left\{ {x_q^*,{f^*}} \right\} = \mathop {{\rm{argmax}}}\limits_{{x_q},f} \mathop \sum \limits_{x_i \in L \cup {x_q}} \mathop \sum \limits_{k = 1}^C MCC\left( {{y_{ik}},{f_k}\left( x \right)} \right) \\ - \lambda \mathop \sum \limits_{k = 1}^C ||{f_k}||_{\cal H}^2 + {\rm{\beta MCC}}\left( {{\rm{L}} \cup {x_q},{\rm{U}}/{x_q}}, y_L,\widehat{y_{U}} \right)
\end{split}
\end{equation}
where $\mathcal{H}$ is a reproducing kernel Hilbert space and $||f_k||^2_{\cal H}$ is used to constrain the complexity of the classifier. $MCC(\cdot)$ is the MCC loss function. $\widehat{y_U}$ is the labels set for all the unlabeled data, which is calculated by the classifiers $f_k, k=1,...,C$. However, there is a problem in solving (\ref{formula28}) that: the labels of $x_q$ are unknown. Our goal is to find the optimal $x_q$ and $f$ with (\ref{formula28}), and the labels of ${x_q}$ are assigned by oracle after query. Therefore, the labels of ${x_q}$ should be precise before query. We replace the precise labels of $x_q$ with pseudo labels to solve (\ref{formula28}), and obtain the following problem:
\begin{equation}\label{formula29}
\begin{split}
\mathop {{\rm{argmax}}}\limits_{{x_q},f} \mathop \sum \limits_{x_i \in L} \mathop \sum \limits_{k = 1}^C MCC\left( {{y_{ik}},{f_k}\left( x_i \right)} \right) \\
+ \mathop \sum \limits_{k = 1}^C MCC\left( {\widehat {{y_{qk}}},{f_{k}}\left( x_q \right)} \right) - \lambda \mathop \sum \limits_{k = 1}^C {||f_k||}_{\cal H}^2 \\+ {\rm{\beta MCC}}\left( {{\rm{L}} \cup {x_q},{\rm{U}}/{x_q},{y_L},\widehat {{y_U}}} \right)
\end{split}
\end{equation}
where $\widehat {{y_{qk}}}$ is the $k^{th}$ pseudo label for $x_q$. It belongs to $\{1, -1\}$. If $x_q$ contains the $k^{th}$ label, $\widehat {{y_{qk}}}$ is equal to 1, otherwise, $\widehat {{y_{qk}}}$ is equal to -1. In (\ref{formula29}), the first three terms correspond to the regularized risk for all the labeled samples after query, which carries the uncertain information embedded in the current classifier. We call them the uncertain part. Meanwhile, in the last term, the unlabeled data are also embedded in the current classifier to enhance the uncertain part. However, the function of the last term is not just to enhance the uncertain information. The main function of the last term is to describe the distribution difference between the labeled samples after query and all the available samples, which captures the representative information embedded in the labeled samples. $\beta$ balances the uncertain and representative information in the formulation. In the remaining part of this section, we will analyze this objective function in a specific form and propose a practical algorithm to solve the optimization problem.

\subsection{Uncertainty based on MCC}
 Minimum margin is the most popular and direct approach to measure the uncertainty of the unlabeled sample by its position to the boundary. Let $f^*$ be the classifier that is trained by the labeled samples. The sample $x_q$ that we want to query in the unlabeled data based on the margin can be found as follows:
\begin{equation}\label{formula4}
{x_q} = \mathop {\arg \min }\limits_{{x_i} \in U} \left| {{f^*}\left( {{x_i}} \right)} \right|
\end{equation}

Generally, with the labeled samples, we can find a classification model $f^*$ for a binary class problem in a supervised learning approach with the following objective function
\begin{equation}\label{formula5}
{f^*} = \mathop {\arg \min }\limits_{f \in {\mathcal{H}}} \sum\nolimits_{{x_i} \in L} {\ell \left( {{Y_i},f\left( {{x_i}} \right)} \right)}  + \lambda \left\| f \right\|_{\mathcal{H}}^2
\end{equation}
where $\mathcal{H}$ is a reproducing kernel Hilbert space endowed with kernel function $K(\cdot); \ell(\cdot)$ is the loss function, and $Y_i$ belongs to $\{1, -1\}$. Following the works of \cite{s3,s32}, Proposition \ref{Proposition1} connects the margin based query selection with the min-max formulation of active learning.
\begin{prop}
The criterion of the minimum margin to find a desirable sample $x\in U$ can be written as
\begin{displaymath}
\begin{split}
{x} = \mathop {\arg \min }\limits_{{x_j} \in U} \mathop {\max }\limits_{\hat{Y_j} \in \{ 1, - 1\} } \mathop {\min }\limits_{f \in {\mathcal{H}}} \sum\nolimits_{{x_i} \in L} {\ell \left( {{Y_i},f\left( {{x_i}} \right)} \right)}  + \lambda \left\| f \right\|_{\mathcal{H}}^2 \\
+ \ell \left( {\hat{Y_j},f\left( {{x_j}} \right)} \right)
\end{split}
\end{displaymath}
where $\hat{Y_j}\in \{1,-1\}$ is the pseudo label for the sample ${x_j\in U}$.
\label{Proposition1}
\end{prop}

In previous works, the loss function is usually adopted with quadratic loss for MSE. But it is not robust for the occasion of outliers. To overcome this problem, we introduce the MCC as the loss function, given by
\begin{equation}\label{formula6}
\begin{split}
\mathop {\arg \max }\limits_{{x_j} \in U} \mathop {\max }\limits_{\hat{Y_j} \in \{ 1, - 1\} } \mathop {\max }\limits_{f \in {\mathcal{H}}} \sum\limits_{{x_i} \in L} {\exp \left( { - \frac{{{{\left\| {{Y_i} - f\left( {{x_i}} \right)} \right\|}^2}}}{{2{\sigma ^2}}}} \right)}  \\
- \lambda \left\| f \right\|_{\mathcal{H}}^2 + \exp \left( { - \frac{{{{\left\| {\hat{Y_j} - f\left( {{x_j}} \right)} \right\|}^2}}}{{2{\sigma ^2}}}} \right)
\end{split}
\end{equation}
where $\sigma$ is the kernel width. Following the Proposition \ref{Proposition1}, we can observe that the objective function (\ref{formula4}) is equal to the objective function (\ref{formula6}). To solve (\ref{formula6}), since $\hat{Y_j}\in \{1, -1\}$, we define $\hat{Y_j}=-sign(f(x_j))$, and optimize the worst case of (\ref{formula6}) for selection. The objective function becomes
\begin{equation}\label{formula7}
\begin{split}
\mathop {\arg \max }\limits_{{x_j} \in U,f \in {\mathcal{H}}} \sum\limits_{{x_i} \in L} {\exp \left( { - \frac{{{{\left\| {{Y_i} - f\left( {{x_i}} \right)} \right\|}^2}}}{{2{\sigma ^2}}}} \right)}  - \lambda \left\| f \right\|_{\mathcal{H}}^2\\
 + \exp \left( { - \frac{{\left( {1 + 2\left| {f\left( {{x_j}} \right)} \right| + f{{\left( {{x_j}} \right)}^2}} \right)}}{{2{\sigma ^2}}}} \right)
\end{split}
\end{equation}

In our work, we extend multi-label classification as several binary classification problems with label correlation. For the convenience of presentation, we consider the simple case by learning one classifier for each label independently. Then, we use the summation of each binary classifier as the minimum margin in multi-label learning, presented by
\begin{equation}\label{formula8}
\mathop {\arg \min }\limits_{{x_j} \in U} \sum\limits_{k = 1}^C {\left| {f_k^*\left( {{x_j}} \right)} \right|}
\end{equation}
where $f_k^*$ is the binary classifier between the $k^{th}$ label and the other labels, $k=\{1,2,...,C\}$. Then the objective function of multi-label learning task based on MCC can be formalized as:
\begin{equation}\label{formula9}
\begin{split}
\mathop {\arg \max }\limits_{{x_j} \in U,{f_k} \in {\mathcal{H}}:k = \left\{ {1,2,..C} \right\}} \sum\limits_{{x_i} \in L} {\sum\limits_{k = 1}^C {\exp \left( { - \frac{{{{\left\| {{y_{ik}} - {f_k}\left( {{x_i}} \right)} \right\|}^2}}}{{2{\sigma ^2}}}} \right)} }  \\
- \lambda \sum\limits_{k = 1}^C {\left\| {{f_k}} \right\|_{\mathcal{H}}^2}  + \sum\limits_{k = 1}^C {\exp \left( { - \frac{{\left( {1 + 2\left| {{f_k}\left( {{x_j}} \right)} \right| + {f_k}{{\left( {{x_j}} \right)}^2}} \right)}}{{2{\sigma ^2}}}} \right)}
\end{split}
\end{equation}
\subsection{Representativeness based on MCC}
Since the labeled samples in $L$ are very limited, it is very important to utilize the unlabeled data to enhance the performance of active learning. However, the labels of the unlabeled data are unknown, and it is hard to add the unlabeled data in a supervised model. To enhance the uncertain information, we merge the representative information into the uncertain information by prediction labels of the unlabeled data. The current similarity measurement is based on the instance features, and it cannot use the unlabeled data to enhance the uncertain information. To overcome this problem, and consider the outlier labels' influence, we take the prediction labels of the unlabeled data into consideration for similarity measurement. We define a novel consistency with labels similarity and features similarity of two instances based on MCC as:
\begin{equation}\label{formula10}
s\left( {\left( {{x_i},{y_i}} \right),\left( {{x_j},{y_j}} \right)} \right) = \exp \left( { - \frac{{\left\| {{y_i} - {y_j}} \right\|_2^2}}{{2{\sigma ^2}}}} \right){w_{ij}}
\end{equation}
where $w_{ij}$ is the similarity between two samples with kernel function $w_{ij}=exp(-||x_i-x_j||^2/{2\sigma^2})$. Let $\boldsymbol{S} = [s_{ij}]^{u \times u}$ denote the symmetric similarity matrix for the unlabeled data, where $s_{ij}$ is the consistency between $x_i$ and $x_j$. We can collect the consistency in a matrix as:
\begin{equation}\label{formula11}
{\boldsymbol{S}_{u \times u}} = \left[ {\begin{array}{*{20}{c}}
{\boldsymbol{s}_1^T}\\
{\boldsymbol{s}_2^T}\\
 \vdots \\
{\boldsymbol{s}_u^T}
\end{array}} \right] = \left[ {\begin{array}{*{20}{c}}
{{s_{11}}}&{{s_{12}}}& \ldots &{{s_{1u}}}\\
{{s_{21}}}&{{s_{22}}}& \ldots &{{s_{2u}}}\\
 \vdots & \vdots & \ddots & \vdots \\
{{s_{u1}}}&{{s_{u2}}}& \ldots &{{s_{uu}}}
\end{array}} \right] \rightarrow {R^{u \times u}}
\end{equation}

With such a consistency matrix, the representativeness is to find the sample that can well represent the unlabeled data set. To do so, \cite{s4} proposed a convex optimization framework by introducing variables $p_{ij}\in [0,1]$ which indicates the probability that $x_i$ represents $x_j$, and collected it with a matrix
\begin{equation}\label{formula12}
{\boldsymbol{P}_{u \times u}} = \left[ {\begin{array}{*{20}{c}}
{\boldsymbol{p}_1^T}\\
{\boldsymbol{p}_2^T}\\
 \vdots \\
{\boldsymbol{p}_u^T}
\end{array}} \right] = \left[ {\begin{array}{*{20}{c}}
{{p_{11}}}&{{p_{12}}}& \ldots &{{p_{1u}}}\\
{{p_{21}}}&{{p_{22}}}& \ldots &{{p_{2u}}}\\
 \vdots & \vdots & \ddots & \vdots \\
{{p_{u1}}}&{{p_{u2}}}& \ldots &{{p_{uu}}}
\end{array}} \right] \rightarrow {R^{u \times u}}
\end{equation}

In our consistency measurement based on MCC, if $x_i$ is very similar to the point $x_j$, and it is not similar to the point $x_t$, there will be $s_{ij}\gg s_{it}$. Such a consistency measurement has already made the difference between representatives and non-representatives large. Therefore, we define $\boldsymbol{p}_i$ as $\boldsymbol{1}_u$, if $x_i$ is the representative one, where $\boldsymbol{1}_u$ is a vector with $u$ length and all the entries in $\boldsymbol{1}_u$ are 1; otherwise $\boldsymbol{p}_i$ is $\boldsymbol{0}_u$, where $\boldsymbol{0}_u$ is a vector with $u$ length and all the entries in $\boldsymbol{0}_u$ are 0. Obviously, it is the summation of consistency between $x_q$ and the unlabeled data. Hence, we can collect the similarities of the query sample and the unlabeled data as:
\begin{equation}\label{formula13}
\sum\limits_{{x_i},{x_j} \in U} {{s_{ij}}{p_{ij}}}  = \sum\limits_{{x_q};{x_j} \in U} {{s_{qj}}{p_{qj}}}
\end{equation}

Similarly, let $\boldsymbol{d} = [d_{ij}]^{u\times l}$ and $\boldsymbol{z} = [z_{ij}]^{u\times l}$ be the consistency matrix and probability between the unlabeled data and the labeled data respectively. The similarities of the query sample and the labeled data can be collected as follows:
\begin{equation}\label{formula14}
\sum\limits_{{x_i} \in L,{x_j} \in U} {{d_{ij}}{z_{ij}}}  = \sum\limits_{{x_q};{x_j} \in L} {{d_{qj}}{z_{qj}}}
\end{equation}

To query a desirable representative sample, which can not only represent the unlabeled data but also has no overlap information with the labeled data, the description of the representative sample on the unlabeled data and labeled data is in contrast. Hence, we maximize the difference of (13) and (14) to measure the representativeness as:
\begin{equation}\label{formula15}
\mathop {\max }\limits_{{x_i} \in U} \sum\limits_{{x_j} \in U} {{s_{ij}}{p_{ij}}}  - \sum\limits_{{x_j} \in L} {{d_{ij}}{z_{ij}}}
\end{equation}

Since there is a large difference between the number of unlabeled data and labeled data, we use the expectation operator and a tradeoff parameter to surrogate them
\begin{equation}\label{formula16}
\begin{split}
&\mathop {\max }\limits_{{x_i} \in U} E\left[ {{x_i} \in U,U} \right] - \beta_0 E\left[ {{x_i} \in U,L} \right] \\
& = \mathop {\max }\limits_{{x_i} \in U} \left( {\frac{1}{u}\sum\limits_{{x_j} \in U} {{s_{ij}}{p_{ij}}} } \right) - \beta_0 \left( {\frac{1}{l}\sum\limits_{{x_j} \in L} {{d_{ij}}{z_{ij}}} } \right)
\end{split}
\end{equation}

\subsection{The Proposed Robust Multi-label Active Learning}
To enhance the query information of uncertainty and representativeness, in our approach, we combine them with a tradeoff parameter, and the objective function can be presented as:
\begin{equation}\label{formula17}
\begin{split}
&\mathop {\arg \max }\limits_{{x_j} \in U,{f_k} \in {\mathcal{H}}:k = \left\{ {1,2,..C} \right\}} \sum\limits_{{x_i} \in L} {\sum\limits_{k = 1}^C {\exp \left( { - \frac{{{{\left\| {{y_{ik}} - {f_k}\left( {{x_i}} \right)} \right\|}^2}}}{{2{\sigma ^2}}}} \right)} } \\
& - \lambda \sum\limits_{k = 1}^C {\left\| {{f_k}} \right\|_{\mathcal{H}}^2}+ \sum\limits_{k = 1}^C {\exp \left( { - \frac{{\left( {1 + 2\left| {{f_k}\left( {{x_j}} \right)} \right| + {f_k}{{\left( {{x_j}} \right)}^2}} \right)}}{{2{\sigma ^2}}}} \right)} \\
&+ {\beta _1}E\left[ {{x_j} \in U,U} \right] - {\beta _2}E\left[ {{x_j} \in U,L} \right]
\end{split}
\end{equation}

To merge the representative part into uncertain part, we use the prediction labels of the unlabeled data. Denoting $f(x_j) = [f_1(x_j), f_2(x_j),..., f_C(x_j)]$ as the prediction labels set for the sample $x_j$ in the unlabeled data, the objective function based on MCC can be defined as:
\begin{equation}\label{formula18}
\begin{split}
&\mathop {\arg \max }\limits_{{x_j} \in U,{f_k} \in {\mathcal{H}}:k = \left\{ {1,2,..C} \right\}} \sum\limits_{{x_i} \in L} {\sum\limits_{k = 1}^C {\exp \left( { - \frac{{{{\left\| {{y_{ik}} - {f_k}\left( {{x_i}} \right)} \right\|}^2}}}{{2{\sigma ^2}}}} \right)} }\\
 & - \lambda \sum\limits_{k = 1}^C {\left\| {{f_k}} \right\|_{\mathcal{H}}^2}  + \sum\limits_{k = 1}^C {\exp \left( { - \frac{{\left( {1 + 2\left| {{f_k}\left( {{x_j}} \right)} \right| + {f_k}{{\left( {{x_j}} \right)}^2}} \right)}}{{2{\sigma ^2}}}} \right)} \\
&+ {\beta _1}\left( {\frac{1}{u}} \right)\sum\limits_{{x_j};{x_i} \in U} {\exp \left( { - \frac{{\left\| {f\left( {{x_j}} \right) - f\left( {{x_i}} \right)} \right\|_2^2}}{{2{\sigma ^2}}}} \right){w_{ij}}} \\
 &- {\beta _2}\left( {\frac{1}{l}} \right)\sum\limits_{{x_j};{x_i} \in L} {\exp \left( { - \frac{{\left\| {f\left( {{x_j}} \right) - {y_i}} \right\|_2^2}}{{2{\sigma ^2}}}} \right)} {w_{ij}}
\end{split}
\end{equation}

To query the specific point $x_q$ from the unlabeled data in the objective function (\ref{formula18}), we use the numerical optimization-based techniques. An indicator vector $\boldsymbol{\alpha}$ is introduced, which is a binary vector with $u$ length. Each entry $\alpha_j$ denotes whether the corresponding sample $x_j$ is queried as $x_q$. If $x_j$ is queried as $x_q$, $\alpha_j$ is 1; otherwise, $\alpha_j$ is 0. Then, the optimization can be formulated as:
\begin{equation}\label{formula19}
\begin{split}
&\mathop {\arg \max }\limits_{{x_j} \in U,{f_k} \in {\mathcal{H}}:k = \left\{ {1,2,..C} \right\}} \sum\limits_{{x_i} \in L} {\sum\limits_{k = 1}^C {\exp \left( { - \frac{{{{\left\| {{y_{ik}} - {f_k}\left( {{x_i}} \right)} \right\|}^2}}}{{2{\sigma ^2}}}} \right)} }  \\
&- \lambda \sum\limits_{k = 1}^C {\left\| {{f_k}} \right\|_{\mathcal{H}}^2}  \\
&+ \sum\limits_{{x_j} \in U} {{\alpha _j}\sum\limits_{k = 1}^C {\exp \left( { - \frac{{\left( {1 + 2\left| {{f_k}\left( {{x_j}} \right)} \right| + {f_k}{{\left( {{x_j}} \right)}^2}} \right)}}{{2{\sigma ^2}}}} \right)} } \\
 &+ {\frac{{\beta _1}}{u}}\sum\limits_{{x_j} \in U} {{\alpha _j}\sum\limits_{{x_i} \in U} {\exp \left( { - \frac{{\left\| {f\left( {{x_j}} \right) - f\left( {{x_i}} \right)} \right\|_2^2}}{{2{\sigma ^2}}}} \right){w_{ji}}} } \\
  &- {\frac{{\beta _2}}{l}}\sum\limits_{{x_j} \in U} {{\alpha _j}\sum\limits_{{x_i} \in L} {\exp \left( { - \frac{{\left\| {f\left( {{x_j}} \right) - {y_i}} \right\|_2^2}}{{2{\sigma ^2}}}} \right){w_{ji}}} }
\end{split}
\end{equation}

For a binary classifier, we use a linear regression model in the kernel space as the classifier $f_k(x)=\omega_k^T\Phi(x)$, where $\Phi(x)$ is the feature mapping to the kernel space, and then the labels set $f(x)$ for $x$ with multi-label can be given by
\begin{equation}\label{formula20}
f\left( x \right) = {\left[ {\begin{array}{*{20}{c}}
{{f_1}\left( x \right)}\\
{{f_2}\left( x \right)}\\
 \vdots \\
{{f_C}\left( x \right)}
\end{array}} \right]^T} = {\left[ {\begin{array}{*{20}{c}}
{{\omega _1}}\\
{{\omega _2}}\\
 \vdots \\
{{\omega _C}}
\end{array}} \right]^T}\left[ {\boldsymbol{R} \otimes \Phi \left( x \right)} \right]
\end{equation}
where $\boldsymbol{R}$ is an identity matrix of size $C \times C$, and it can also be the label correlation matrix. $\otimes$ is the kronecker product between matrices. We define $\boldsymbol{\omega}=[\omega_1,\omega_2,...,\omega_C]^T$, and then the multi-label classifier $f(x)$ can be presented by $f(x) = \boldsymbol{\omega}^T[\boldsymbol{R}\otimes\Phi(x)]$. The objective function can be formalized as:
\begin{equation}\label{formula21}
\begin{split}
&\mathop {\arg \max }\limits_{\boldsymbol{\omega} ;{\boldsymbol{\alpha} ^T}\boldsymbol{1}_u = 1:{\alpha _j} \in \left\{ {0,1} \right\}} \sum\limits_{{x_i} \in L} {\sum\limits_{k = 1}^C {\exp \left( { - \frac{{{{\left\| {{y_{ik}} - \omega _k^T\Phi \left( {{x_i}} \right)} \right\|}^2}}}{{2{\sigma ^2}}}} \right)} }\\
 & - \lambda \sum\limits_{k = 1}^C {{{\left\| {{\omega _k}} \right\|}^2}}\\
   &+ \sum\limits_{{x_j} \in U} {{\alpha _j}\sum\limits_{k = 1}^C {\exp \left( { - \frac{{\left( {1 + 2\left| {\omega _k^T\Phi \left( {{x_j}} \right)} \right| + {{\left( {\omega _k^T\Phi \left( {{x_j}} \right)} \right)}^2}} \right)}}{{2{\sigma ^2}}}} \right)} } \\
  &+ \frac{{{\beta _1}}}{u}\sum\limits_{{x_j},{x_i} \in U} { {{{\alpha _j}}} {\exp \left( { - \frac{{\left\| {{\boldsymbol{\omega} ^T}\left[ {\boldsymbol{R} \otimes \left[ {\Phi \left( {{x_i}} \right) - \Phi \left( {{x_j}} \right)} \right]} \right]} \right\|_2^2}}{{2{\sigma ^2}}}} \right){w_{ji}}} }\\
  &- \frac{{{\beta _2}}}{l}\sum\limits_{{x_j} \in U} {\alpha _j\sum\limits_{{x_i} \in L} {\exp \left( { - \frac{{\left\| {{\boldsymbol{\omega} ^T}\left[ {\boldsymbol{R} \otimes \Phi \left( {{x_j}} \right)} \right] - {y_i}} \right\|_2^2}}{{2{\sigma ^2}}}} \right){w_{ji}}} }
\end{split}
\end{equation}
where $\boldsymbol{1}_u$ is a length of vector $u$. We derive an iterative algorithm based on half-quadratic technique with alternating optimization strategy to solve (\ref{formula21}) efficiently\cite{s33}. Based on the theory of convex conjugated functions, we can easily derive the proposition \ref{Proposition2} \cite{s34,s35}.
\begin{prop}
A convex conjugate function $\varphi$ exits to make sure
\begin{displaymath}
g\left( x \right) = \exp \left( { - \frac{{{x^2}}}{{2{\sigma ^2}}}} \right) = \mathop {\max }\limits_{p'} \left( {p'\frac{{{{\left\| x \right\|}^2}}}{{{\sigma ^2}}} - \varphi \left( {p'} \right)} \right)
\end{displaymath}
where $p'$ is the auxiliary variable, and with a fixed $x$, $g(x)$ reaches the maximum value at $p' = -g(x)$.
\label{Proposition2}
\end{prop}
Following the Proposition \ref{Proposition2}, the objective function (\ref{formula21}) can be formulated as:
\begin{equation}\label{formula22}
\begin{split}
&\mathop {\arg \min }\limits_{\boldsymbol{\omega} ;{\boldsymbol{\alpha} ^T}\boldsymbol{1}_u = 1,{\alpha _i} \in \left\{ {0,1} \right\}} \sum\limits_{{x_i} \in L} {\sum\limits_{k = 1}^C {\left[ {{m_{ik}}{{\left\| {{y_{ik}} - \omega _k^T\Phi \left( {{x_i}} \right)} \right\|}^2}} \right]} } \\
 &+ \lambda \sum\limits_{k = 1}^C {{{\left\| {{\omega _k}} \right\|}^2}} \\
 &+ \sum\limits_{{x_j} \in U} {{\alpha _j}\sum\limits_{k = 1}^C {\left[ {{n_{jk}}\left( {1 + 2\left| {\omega _k^T\Phi \left( {{x_j}} \right)} \right| + {{\left( {\omega _k^T\Phi \left( {{x_j}} \right)} \right)}^2}} \right)} \right]} } \\
&-{\beta _1}\sum\limits_{{x_j} \in U} {{\alpha _j}\sum\limits_{{x_i} \in U} {{h_{ji}}\left\| {{\boldsymbol{\omega} ^T}\left[ {\boldsymbol{R} \otimes \left( {\Phi \left( {{x_j}} \right) - \Phi \left( {{x_i}} \right)} \right)} \right]} \right\|_2^2{w_{ji}}} }\\
 &+ {\beta _2}\sum\limits_{{x_j} \in U} {{\alpha _j}\sum\limits_{{x_i} \in L} {{v_{ji}}\left\| {{\boldsymbol{\omega} ^T}\left[ {\boldsymbol{R} \otimes \Phi \left( {{x_j}} \right)} \right] - {y_i}} \right\|_2^2{w_{ji}}} }
\end{split}
\end{equation}
where $m_{ik}, n_{jk}, h_{ji}$, and $v_{ji}$ with $x_i, x_j \in U$ are the auxiliary variables, with
\begin{displaymath}
\begin{split}
&{m_{ik}} = \exp \left( { - \frac{{{{\left\| {{y_{ik}} - \omega _k^T\Phi \left( {{x_i}} \right)} \right\|}^2}}}{{2{\sigma ^2}}}} \right),{x_i} \in L,{y_{ik}} \in {y_i}\\
&{n_{jk}} = \exp \left( { - \frac{{\left( {1 + 2\left| {\omega _k^T\Phi \left( {{x_j}} \right)} \right| + {{\left( {\omega _k^T\Phi \left( {{x_j}} \right)} \right)}^2}} \right)}}{{2{\sigma ^2}}}} \right)\\
&{h_{ji}} = \frac{1}{u}\exp \left( { - \frac{{\left\| {{\boldsymbol{\omega} ^T}[\boldsymbol{R}\otimes\Phi \left( {{x_j}} \right)] - {\boldsymbol{\omega} ^T}[\boldsymbol{R}\otimes\Phi \left( {{x_i}} \right)]} \right\|_2^2}}{{2{\sigma ^2}}}} \right)\\
&{v_{ji}} = \frac{1}{l}\exp \left( { - \frac{{\left\| {{\boldsymbol{\omega} ^T}[\boldsymbol{R}\otimes\Phi \left( {{x_j}} \right)] - {y_i}} \right\|_2^2}}{{2{\sigma ^2}}}} \right),{y_i} \in y_L
\end{split}
\end{displaymath}

The objective function (\ref{formula22}) can be solved by the alternating optimization strategy.

Firstly, $\boldsymbol{\alpha}$ is fixed, and the objective function (\ref{formula22}) is to find the optimal classifier $\boldsymbol{\omega}$. It can be solved by the alternating direction method of multipliers (ADMM)\cite{s36}.

Secondly, $\boldsymbol{\omega}$ obtained in the first step is fixed, and the objective function (\ref{formula22}) becomes
\begin{equation}\label{formula23}
\mathop {\arg \max }\limits_{{\boldsymbol{\alpha} ^T}\boldsymbol{1}_u = 1:{\alpha _i} \in \left\{ {0,1} \right\}} {\boldsymbol{\alpha} ^T}\boldsymbol{a} + {\beta _1}{\boldsymbol{\alpha} ^T}\boldsymbol{b} - {\beta _2}{\boldsymbol{\alpha} ^T}\boldsymbol{c}
\end{equation}
where
\begin{displaymath}
\begin{split}
&{a_j} = \sum\limits_{k = 1}^C {\exp \left( { - \frac{{\left( {1 + 2\left| {\omega _k^T\Phi \left( {{x_j}} \right)} \right| + {{\left( {\omega _k^T\Phi \left( {{x_j}} \right)} \right)}^2}} \right)}}{{2{\sigma ^2}}}} \right)}\\
&{b_j} = \frac{1}{u}\sum\limits_{{x_i} \in U} {\exp \left( { - \frac{{\left\| {{\boldsymbol{\omega} ^T}\left[ {\boldsymbol{R} \otimes \left( {\Phi \left( {{x_j}} \right) - \Phi \left( {{x_i}} \right)} \right)} \right]} \right\|_2^2}}{{2{\sigma ^2}}}} \right){w_{ji}}}\\
&{c_j} = \frac{1}{l}\sum\limits_{{x_i} \in L} {\exp \left( { - \frac{{\left\| {\boldsymbol{\omega} ^T[\boldsymbol{R}\otimes\Phi \left( {{x_j}} \right)] - {y_i}} \right\|_2^2}}{{2{\sigma ^2}}}} \right){w_{ji}}}
\end{split}
\end{displaymath}

To solve (\ref{formula23}), as in \cite{s20}, we relax $\alpha_j$ to a continuous range [0, 1]. Thus, the $\boldsymbol{\alpha}$ can be solved with a linear program. The sample corresponding to the largest value in $\boldsymbol{\alpha}$ will be queried as $x_q$.
\subsection{The solution}
In this part, we will discuss the details of the algorithm to solve the objective function (\ref{formula21}). We solve it with alternative strategy in two steps.
Firstly, $\boldsymbol{\alpha}$ is fixed. In this step, the classifier is adopted with kernel form, and we use $\omega_k = \sum\limits_{x_i \in L} \theta_{ki}\Phi (x_i)$ to learn $\theta_k$ for each classifier, where $\theta_k = [\theta_{k1}, \theta_{k2},..., \theta_{kl}]^T$. We define $\theta = [\theta_1, \theta_2,..., \theta_C]^T$, and learn $\theta$ from the following formulation
\begin{equation}\label{formula24}
\begin{aligned}
 &\mathop {\arg \min }\limits_\theta  \sum\limits_{{x_i} \in L} {\sum\limits_{k = 1}^C {\left[ {{m_{ik}}{{\left\| {{y_{ik}} - \theta _k^T{K_L}\left( {{x_i}} \right)} \right\|}^2}} \right]} } \\
&+ \lambda {\theta ^T}\left( {\boldsymbol{R} \otimes {K_{LL}}} \right)\theta\\
 &+ \sum\limits_{k = 1}^C {\left[ {{n_{qk}}\left( {1 + 2\left| {\theta _k^T{K_L}\left( {{x_q}} \right)} \right| + {{\left( {\theta _k^T{K_L}\left( {{x_q}} \right)} \right)}^2}} \right)} \right]}\\
 &+ {\beta _1}\left( {\frac{1}{u}} \right)\sum\limits_{{x_i} \in U} {{h_{qi}}\left\| {{\theta ^T}\left[ {\boldsymbol{R} \otimes \left( {{K_L}\left( {{x_i}} \right) - {K_L}\left( {{x_q}} \right)} \right)} \right]} \right\|_2^2{w_{qi}}}\\
 &- {\beta _2}\left( {\frac{1}{l}} \right)\sum\limits_{{x_i} \in L} {{v_{qi}}\left\| {{\theta ^T}\left( {\boldsymbol{R} \otimes {K_L}\left( {{x_q}} \right)} \right) - {y_i}} \right\|_2^2{w_{qi}}}
\end{aligned}
\end{equation}

As stated above, $\boldsymbol{R}$ is an identify matrix. Define $m_i = [m_{i1}, m_{i2},..., m_{iC}], M = [m_1, m_2,...,m_l] \rightarrow R^{l \times C}, N = [n_{q1},n_{q2},...,n_{qC}]. v_{qi}^* = v_{qi}*w_{qi}, h_{qi}^* = h_{qi}*w_{qi}, v^* = [v_{q1}^*, v_{q2}^*, v_{q3}^*,..., v_{ql}^*], h^*= [h_{q1}^*, h_{q2}^*, h_{q3}^*,..., h_{qu}^*]$. Meanwhile, an auxiliary variable $e_k =\theta_k^TK_L(x_q)$ is introduced. Then, the objective function becomes
\begin{equation}\label{formula25}
\begin{aligned}
&\mathop {\arg \min }\limits_\theta  \left( {vec{{(y_L)}^T} - {\theta ^T}\left( {\boldsymbol{R} \otimes {K_{LL}}} \right)} \right)diag(vec(M))\\
&{{\left( {vec{{(y_L)}^T}- {\theta ^T}\left( {\boldsymbol{R} \otimes {K_{LL}}} \right)} \right)}^T} \\
 &+ \lambda {\theta ^T}\left( {\boldsymbol{R} \otimes {K_{LL}}} \right)\theta  + ediag(N){e^T} + 2\left| {diag(N)e} \right|\\
 & +  {\frac{\beta_1}{u}} {\theta ^T}\left( {\boldsymbol{R} \otimes \left( {{K_{LU}}diag\left( {{h^*}} \right){{\left( {{\textbf{1}}_u^T \otimes {K_L}\left( {{x_q}} \right)} \right)}^T}} \right)} \right)\theta \\
& -  {\frac{\beta_2}{l}} \left[ {{\theta ^T}\left( {\boldsymbol{R} \otimes {K_{LL}}} \right) - vec\left( y_L \right)} \right]diag\left( {vec\left( {{\textbf{1}}_l^T \otimes {v^*}} \right)} \right)\\
& {\left[ {{\theta ^T}\left( {\boldsymbol{R} \otimes {K_{LL}}} \right) - vec\left( y_L \right)} \right]^T}\\
&s.t.~~ {e_k} = \theta _k^T{K_L}\left( {{x_q}} \right),\forall {x_q} \in U,k = 1,2,...,C
\end{aligned}
\end{equation}
where $\boldsymbol{1}_u$ and $\boldsymbol{1}_l$ are length of vectors $u$ and $l$ respectively, with all the entries being 1. $y_L$ is the labels matrix for the labeled data. $vec(\cdot)$ is the function to convert a matrix to a vector along the column. The augmented Lagrangian function is given by
\begin{equation}\label{formula26}
\begin{aligned}
&{L_\rho }  =  \left( {vec{{(y_{L})}^T} - {\theta ^T}\left( {\boldsymbol{R} \otimes {K_{LL}}} \right)} \right)diag(vec(M))\\
&{{\left( {vec{{(y_L)}^T} - {\theta ^T}\left( {\boldsymbol{R} \otimes {K_{LL}}} \right)} \right)}^T} \\
& + \lambda {\theta ^T}\left( {\boldsymbol{R} \otimes {K_{LL}}} \right)\theta  + ediag(N){e^T} + 2\left| {diag(N)e} \right|\\
& + \left( {\frac{\beta _1}{u}} \right){\theta ^T}\left( {\boldsymbol{R} \otimes \left( {{K_{LU}}diag\left( {{h^*}} \right){{\left( {{\textbf{1}}_u^T \otimes {K_L}\left( {{x_q}} \right)} \right)}^T}} \right)} \right)\theta \\
&  - \left( {\frac{\beta _2}{l}} \right)\left[ {{\theta ^T}\left( {\boldsymbol{R} \otimes {K_{LL}}} \right) - vec\left( y_L \right)} \right]diag\left( {vec\left( {{\textbf{1}}_l^T \otimes {v^*}} \right)} \right)\\
&  {\left[ {{\theta ^T}\left( {\boldsymbol{R} \otimes {K_{LL}}} \right) - vec\left( y_L \right)} \right]^T}+ \left( {e - {\theta ^T}\left( {\boldsymbol{R} \otimes {K_L}\left( {{x_q}} \right)} \right)} \right){\eta ^T} \\
 &+ \frac{\rho }{2}\left\| {e - {\theta ^T}\left( {\boldsymbol{R} \otimes {K_L}\left( {{x_q}} \right)} \right)} \right\|_2^2
\end{aligned}
\end{equation}
The updating rules are as follows:
\begin{displaymath}
\begin{aligned}
&{\theta ^{t + 1}} = {B^{ - 1}}{r^T}, where\\
&B= \left( {\boldsymbol{R} \otimes {K_{LL}}} \right)diag({M^t}){\left( {\boldsymbol{R} \otimes {K_{LL}}} \right)^T} + \lambda {\left( {\boldsymbol{R} \otimes {K_{LL}}} \right)^2} \\
& + \frac{\rho }{2}\left( {\boldsymbol{R},{K_L}\left( {{x_q}} \right){K_L}{{\left( {{x_q}} \right)}^T}} \right)\\
& + {\beta _1}\left( {\frac{1}{u}} \right)\left( {\boldsymbol{R} \otimes \left( {{K_{LU}}diag\left( {{h^{*t}}} \right){{\left( {{\rm{1}}_u^T \otimes {K_L}\left( {{x_q}} \right)} \right)}^T}} \right)} \right)\\
& - {\beta _2}\left( {\frac{1}{l}} \right)\left( {\boldsymbol{R} \otimes \left( {{K_{LL}}diag\left( {{v^{*t}}} \right){K_{LL}}} \right)} \right)\\
&r= vec{\left( y_L \right)^T}diag\left( M \right)\left( {\boldsymbol{R} \otimes {K_{LL}}} \right)\\
& + 0.5*\left( {{\eta ^t} + \rho {e^t}} \right){\left( {\boldsymbol{R} \otimes {K_L}\left( {{x_q}} \right)} \right)^T}\\
& - {\beta _2}\left( {\frac{1}{l}} \right)\left[ {vec{{\left( y_L \right)}^T}diag\left( {vec\left( {{\boldsymbol{1}}_l^T \otimes {v^{*t}}} \right)} \right){{\left( {\boldsymbol{R} \otimes {K_{LL}}} \right)}^T}} \right]\\
 &{e^{t + 1}}  = \mathop {\arg \min }\limits_e {e}diag(N){e^T} + 2\left| {diag(N)e} \right| + e{\eta ^t}^T \\
 &+ \frac{\rho }{2}e{e^T} - \rho {\theta ^T}\left( {\boldsymbol{R} \otimes {K_L}\left( {{x_q}} \right)} \right){e^T}\\
& = \arg \min \frac{1}{2}\left\| {A{e^T} - \mu } \right\|_2^2 + 2\left| {diag\left( N \right)e} \right|\\
&where~~ \Upsilon  = diag\left( {2N + \rho } \right),A = {\Upsilon ^{ - 1}},\\
&\mu = \left[ {{\eta ^t} - \rho {\theta ^{tT}}\left( {\boldsymbol{R} \otimes {K_L}\left( {{x_q}} \right)} \right)} \right]A\\
&{\eta ^{t + 1}} = {\eta ^t} + \rho \left[ {e - {\theta ^{\left( {t + 1} \right)T}}\left( {\boldsymbol{R} \otimes {K_L}\left( {{x_q}} \right)} \right)} \right]
\end{aligned}
\end{displaymath}

The problem to solve $e$ is a sparse one, and it can be solved with SPLA toolbox\footnote{http://spams-devel.gforge.inria.fr/downloads.html}. It stops until the convergence condition is satisfied.

In the second step, $\boldsymbol{\omega}$ is fixed to solve $\boldsymbol{\alpha}$ $w.r.t$ $x_q$. As stated above, the objective function is
\begin{equation}\label{formula27}
 \max \limits_{{\boldsymbol{\alpha} ^T}\boldsymbol{1}_u = 1:{\alpha _i} \in \left[ {0,1} \right]} {\boldsymbol{\alpha} ^T}H
\end{equation}
where $H = \boldsymbol{a} + \beta_1\boldsymbol{b} - \beta_2\boldsymbol{c}$. The linear program can be used to solve (\ref{formula27}), and we select the most valuable sample $x_q$ that is corresponding to the largest value in $\boldsymbol{\alpha}$. We summarize our algorithm in the Algorithm 1.

\section{Experiments}
In this section, we present the experimental results to validate the effectiveness of the proposed method on 12 multi-label data sets from Mulan project\footnote{http://mulan.sourceforge.net/datasets-mlc.html}. The characteristics of data sets are introduced in Table I. To demonstrate the superiority of our method, several methods listed as follows are regarded as competitors.
\begin{algorithm}[H]
\caption{The Active Learning Framework for Cold-start Recommendation}
\label{alg:Framwork}
\begin{algorithmic}[1]
\REQUIRE ~~
The labeled data set $L$ and the unlabeled data set $U$, the tradeoff parameters $\beta_1$ and $\beta_2$, and the initial variables and parameters.
\REPEAT
\STATE Fix $\boldsymbol{\alpha}$, and calculate the objective function (\ref{formula25}) with ADMM strategy to obtain the values of $\theta$ (${\emph {w.r.t}}$ $\boldsymbol{\omega}$).
\STATE With the values of $\theta$, calculate the indicator vector $\boldsymbol{\alpha}$ by solving (\ref{formula27}), and select the sample that is corresponding to the largest value in $\boldsymbol{\alpha}$.
\UNTIL {the tolerance is satisfied}
\ENSURE~~
The query index of unlabeled samples.
\end{algorithmic}
\end{algorithm}

\begin{enumerate}
\item	RANDOM is the baseline which randomly selects instances for labeling.
\item AUDI\cite{s29} combines label ranking with threshold learning, and then exploits both uncertainty and diversity in the instance space as well as the label space.
\item	Adaptive\cite{s27} combines the max-margin prediction uncertainty and the label cardinality inconsistency as the criterion for active selection.
\item	QUIRE\cite{s3} provides a systematic way for measuring and combining the informativeness and representativeness of an unlabeled instance by incorporating the correlation among labels.
\item Batchrank\cite{s20} selects the best query with an NP-hard optimization problem based on the mutual information.
\item	RMLAL: Robust Multi-label Active Learning is the proposed method in this paper.
\end{enumerate}

\begin{table}\label{table1}
\centering
\caption{Characteristics of the datasets, including the numbers of the corresponding instance,labels, features and cardinality.}
\begin{tabular}{|c|c|c|c|c|c|} \hline
Dataset&domain &\#instance& \#label& \#feature & \#LC\\ \hline
Corel16k&images&13,766&153&500&2.86\\ \hline
Mediamill&video&43,097&101&120&4.37\\ \hline
Emotions&music&	593	&6&	72	&1.87\\ \hline
Enron&text&1,702&	53	&1,001&	3.38\\ \hline
Image	&images&2,000&	5	&294	&1.24\\ \hline
Medical	&text&978	&45	&1,449	&1.25\\ \hline
Scene	&images&2,407	&6	&294	&1.07\\ \hline
Health	&text&5,000	&32	&612	&1.66\\ \hline
Social	&text&5,000	&39	&1,047	&1.28\\ \hline
Corel5k&images & 5000&374&499&3.52\\ \hline
Genbase	&biology&662&	27	&1,185&	1.25\\ \hline
CAL500&music&502&174&68&26.04\\ \hline
\end{tabular}
\end{table}

\begin{table*}[htb]\label{table2}
\centering
\caption{Win/Tie/Loss counts of our method versus the competitors based on paired t-test at 95 percent significance level.}
\begin{tabular}{|c|c|c|c|c|c|} \hline
Dataset&Vs QUIRE&Vs AUDI&Vs Adaptive&Batchrank&Vs Random\\ \hline
Corel16k&25/0/0&25/0/0&25/0/0&25/0/0&25/0/0\\ \hline
Mediamill&5/16/4&10/12/3&25/0/0&25/0/0&25/0/0\\ \hline
Emotions&25/0/0&25/0/0&25/0/0&25/0/0&25/0/0\\ \hline
Enron&19/5/1&25/0/0&25/0/0&25/0/0&25/0/0\\ \hline
Image&15/10/0&17/8/0&25/0/0&25/0/0&25/0/0\\ \hline
Medical&13/10/2&25/0/0&25/0/0&25/0/0&25/0/0\\ \hline
Scene&15/5/5&25/0/0&25/0/0&25/0/0&25/0/0\\ \hline
Health&13/10/2&18/5/2&25/0/0&25/0/0&25/0/0\\ \hline
Social&25/0/0&25/0/0&25/0/0&25/0/0&25/0/0\\ \hline
Corel5k&25/0/0&25/0/0&25/0/0&25/0/0&25/0/0\\ \hline
Genbase&25/0/0&20/5/0&25/0/0&7/15/3&25/0/0\\ \hline
CAL500&25/0/0&25/0/0&25/0/0&25/0/0&25/0/0\\ \hline
\end{tabular}
\end{table*}
LC is the average number of labels for each instance. We randomly divided each data set into two equal parts. One was regarded as testing data set. For the other part, we randomly selected 4\% as the initial labeled set, and the remaining samples of this part were used as the unlabeled data set. In the compared methods, AUDI and QUIRE query a relevant label-instance pairs at each iteration. We can notice that querying all labels for one instance is equal to query $C$ label-instance pairs. Hence, to achieve a fair comparison, we queried $C$ label-instance pairs as one query instance in AUDI and QUIRE. For the method Batchrank, in the original paper, the tradeoff parameter is set as 1. For a fair comparison, we chose the tradeoff parameter from a candidate set that is the same as the proposed method. The other methods and the parameters were all set as the original papers. For the kernel parameters, we adopted the same values for all methods.

Without loss of generality, we adopted the liblinear\footnote{https://www.csie.ntu.edu.tw/~cjlin/liblinear/} as the classifier for all methods, and evaluated the performance with micro-F1\cite{s37}, which is commonly used as performance measurement in the multi-label learning. Following\cite{s8}, for each data set, we repeated each method for 5 times and reported the average results. We stopped the querying process when 100 iterations were reached and one instance was queried at each iteration.
\begin{figure*}[htbp]\label{fig4}
\begin{center}
\subfigure[Corel16k]{
\epsfig{file = 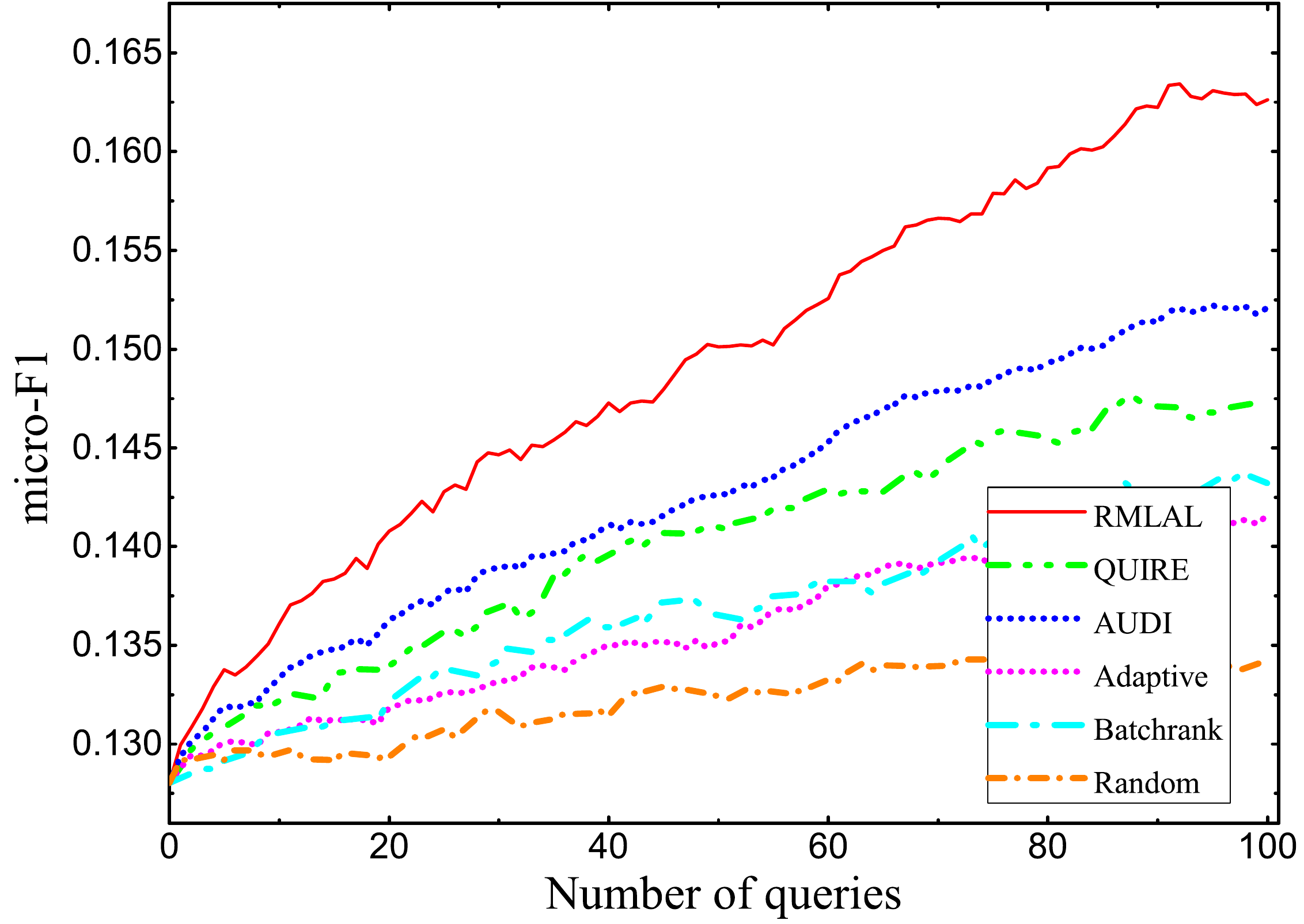, width=2.2 in,height=1.8 in}}
\subfigure[Mediamill]{
\epsfig{file = 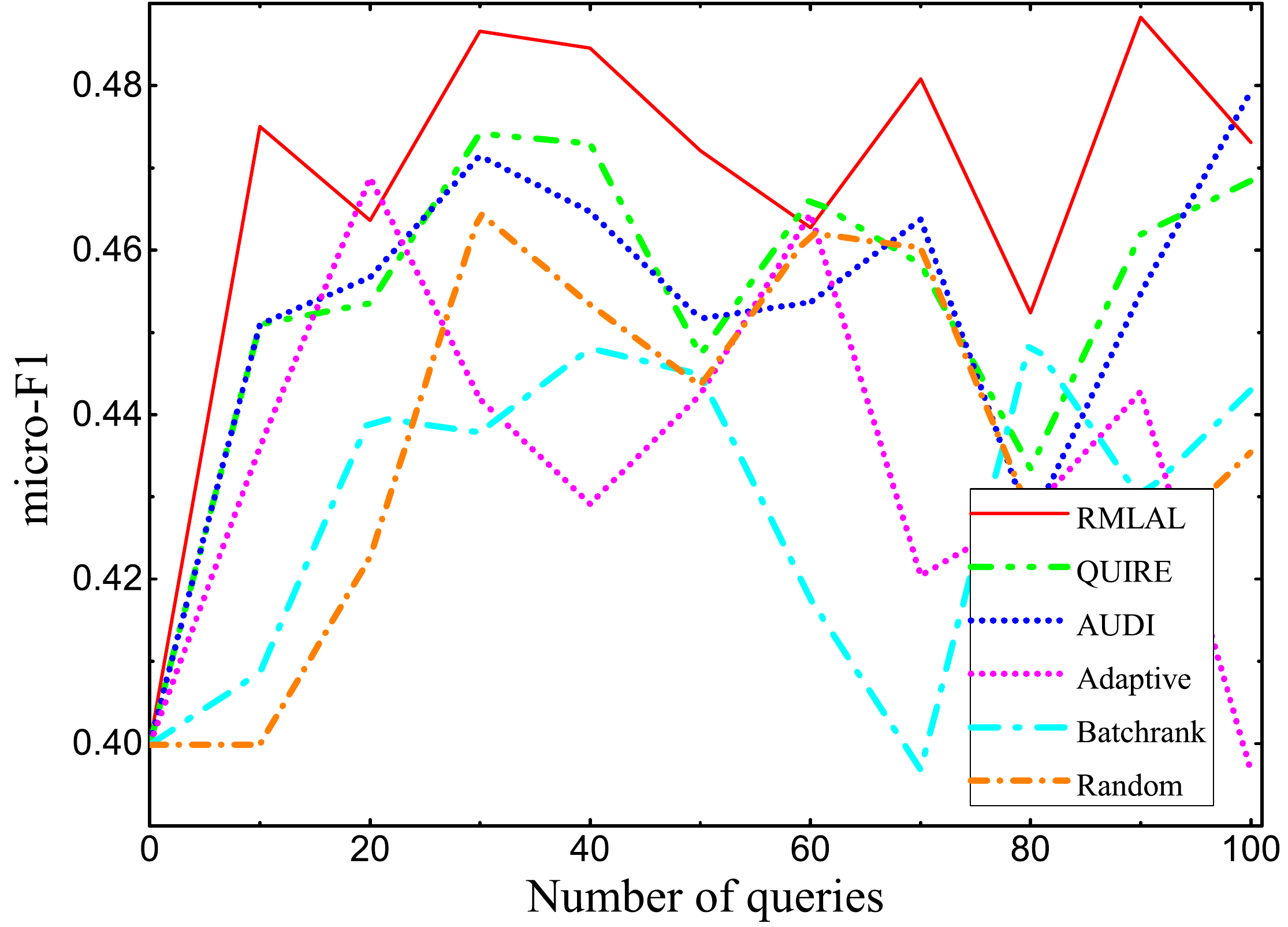, width=2.2 in,height=1.8 in}}
\subfigure[Enron]{
\epsfig{file = 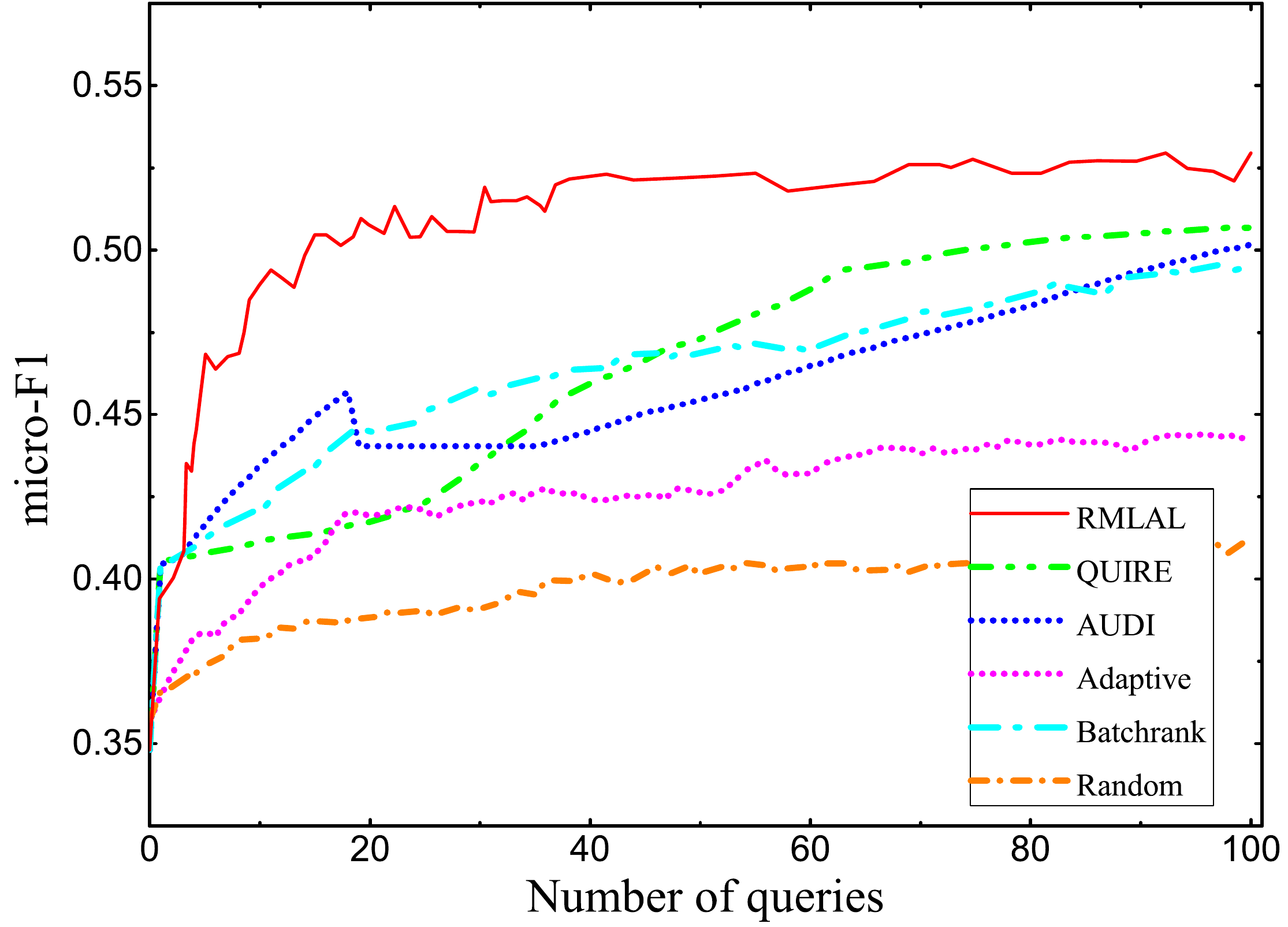, width=2.2 in,height=1.8 in}}
\subfigure[Image]{
\epsfig{file = 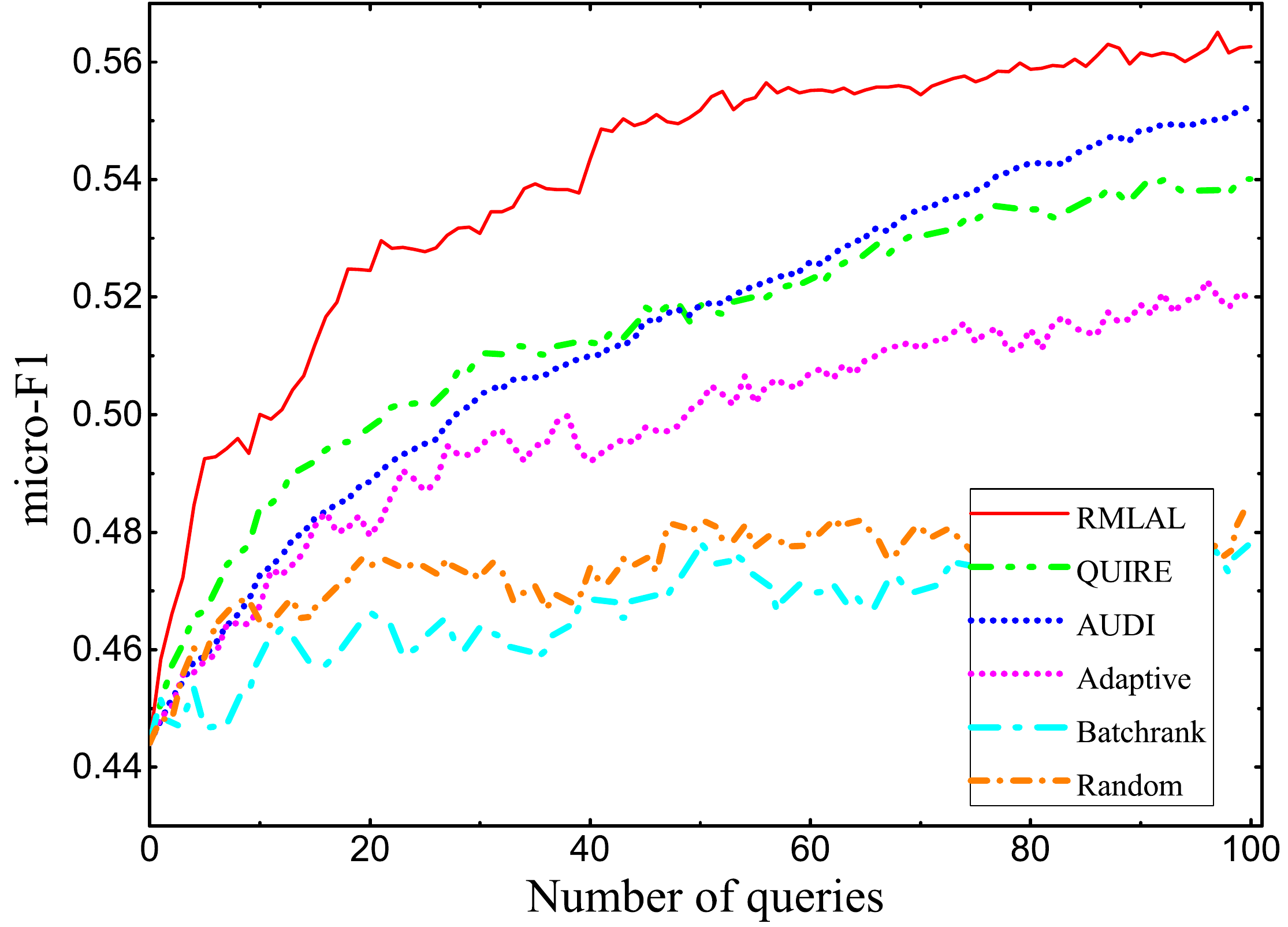, width=2.2 in,height=1.8 in}}
\subfigure[Scene]{
\epsfig{file = 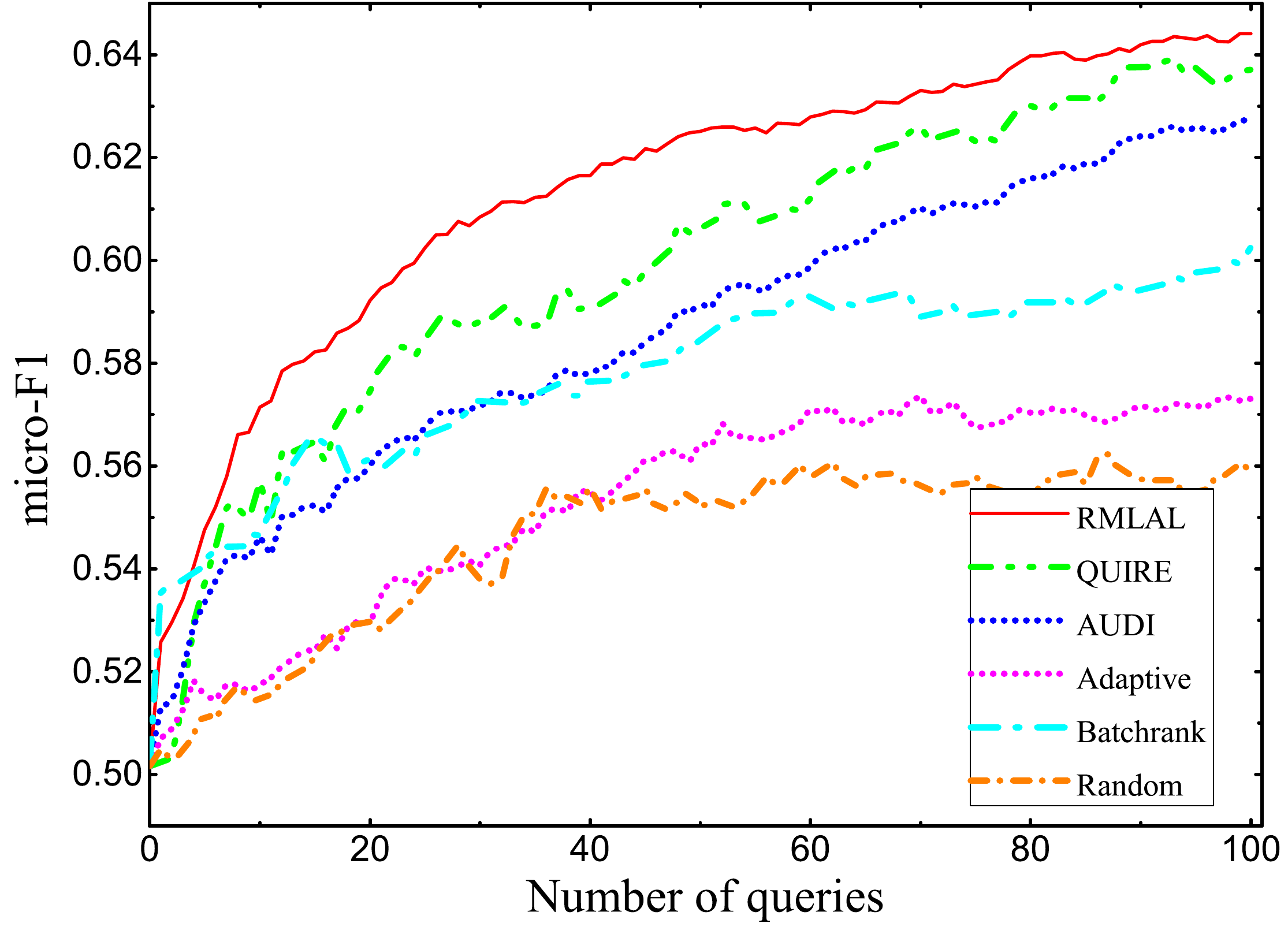, width=2.2 in,height=1.8 in}}
\subfigure[Health]{
\epsfig{file = 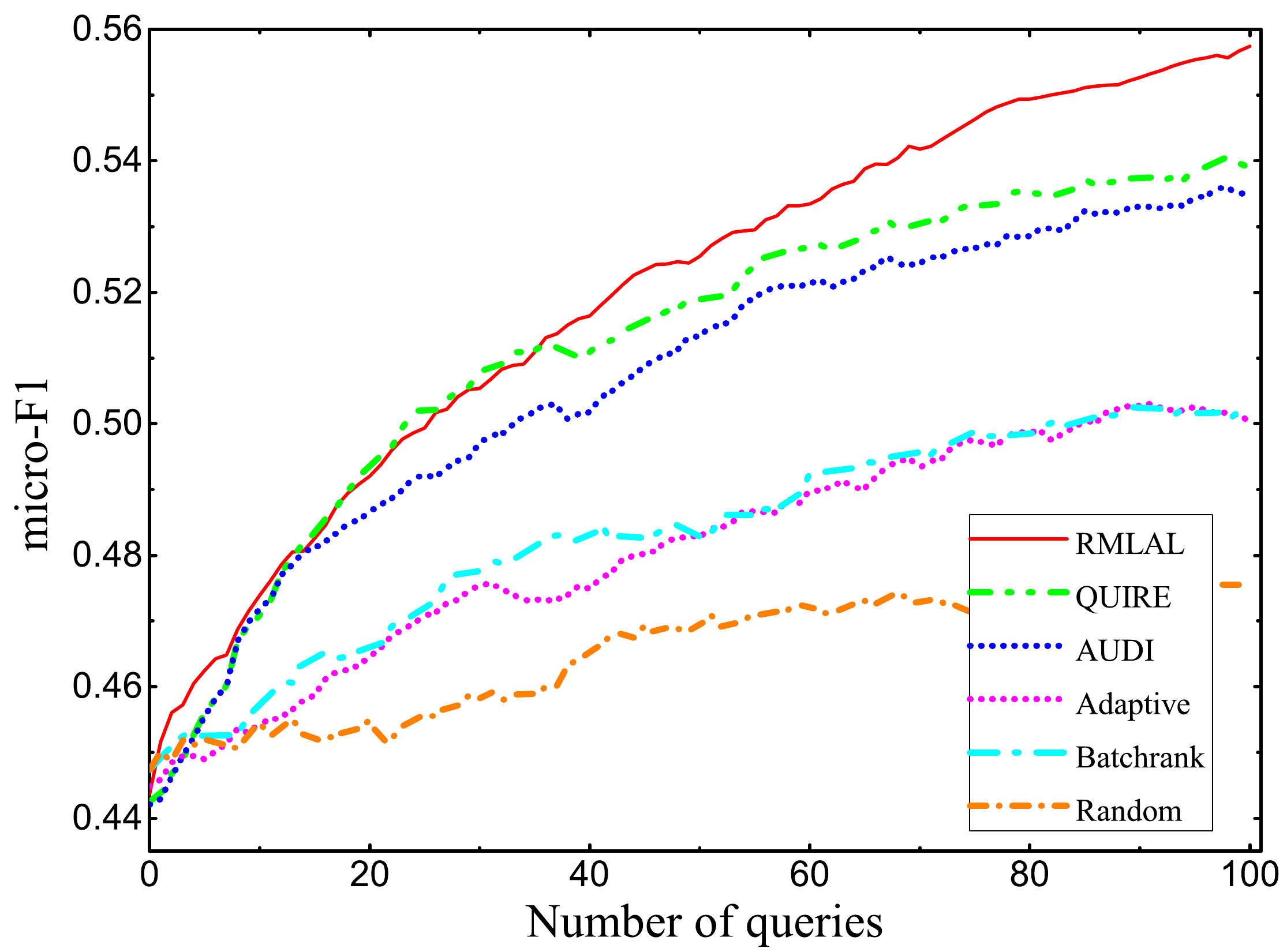, width=2.2 in,height=1.8 in}}
\subfigure[Emotions]{
\epsfig{file = 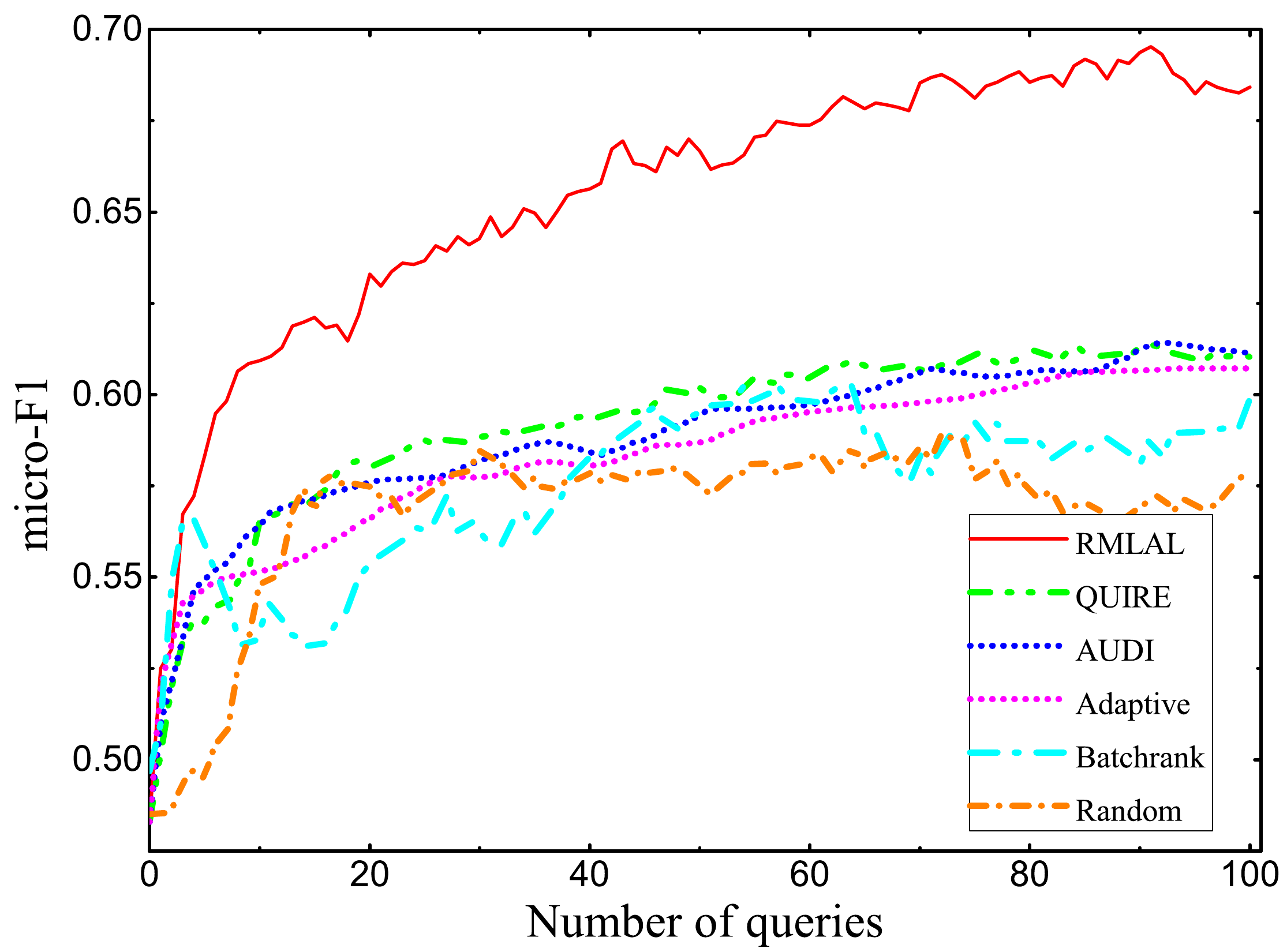, width=2.2 in,height=1.8 in}}
\subfigure[Medical]{
\epsfig{file = 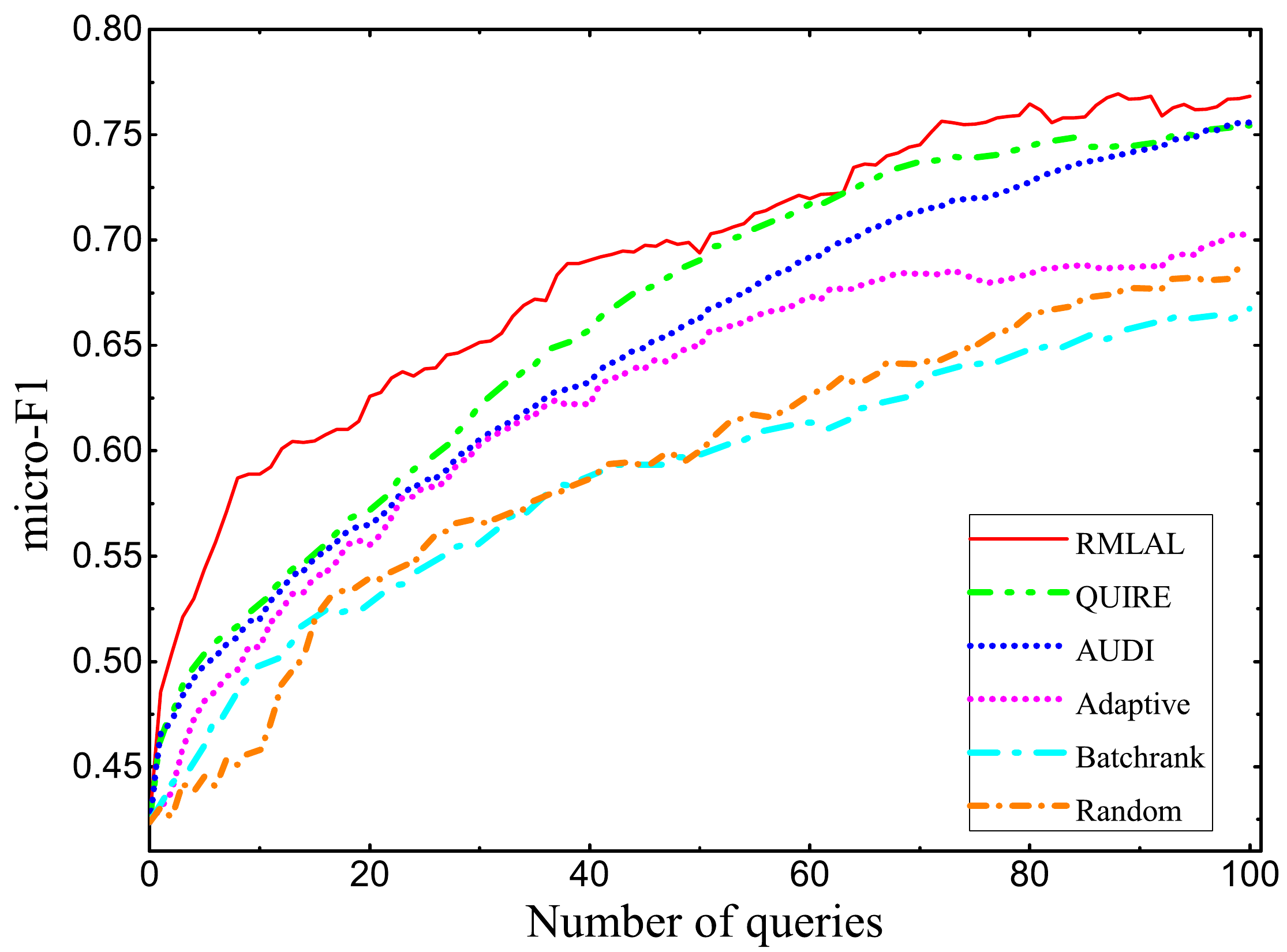, width=2.2 in,height=1.8 in}}
\subfigure[Social]{
\epsfig{file = 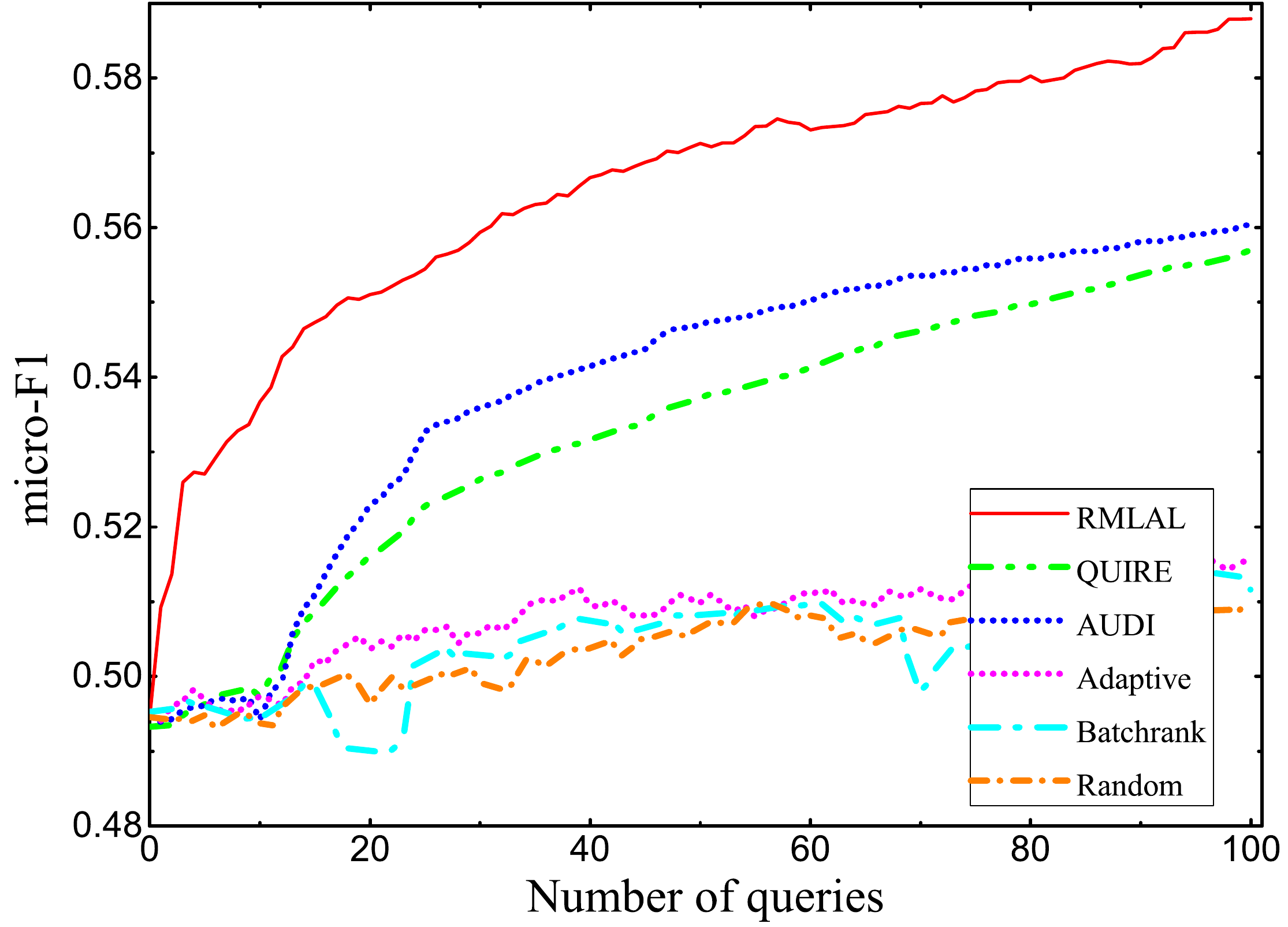, width=2.2 in,height=1.8 in}}
\subfigure[Corel5k]{
\epsfig{file = 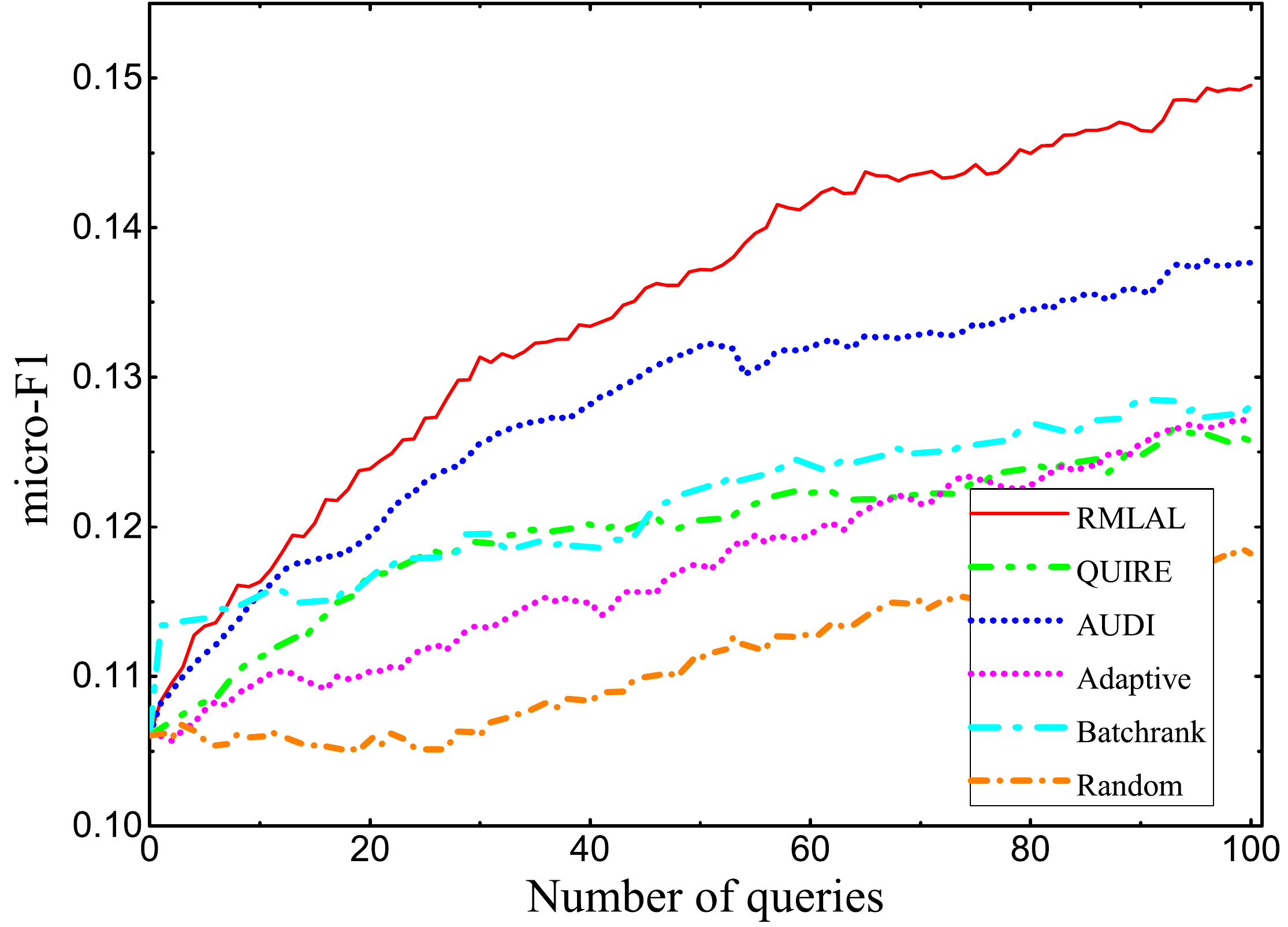, width=2.2 in,height=1.8 in}}
\subfigure[Genbase]{
\epsfig{file = 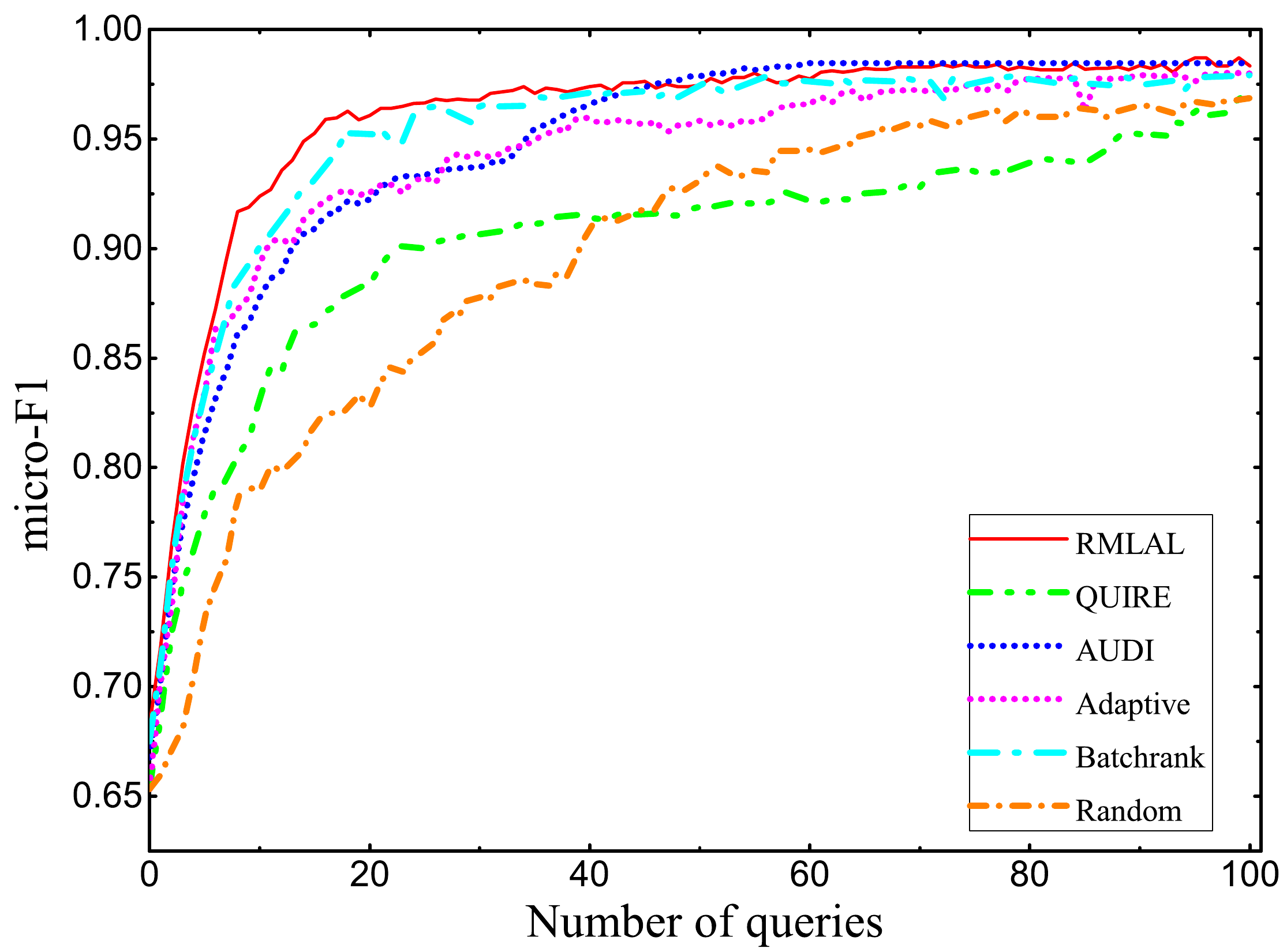, width=2.2 in,height=1.8 in}}
\subfigure[CAL500]{
\epsfig{file = 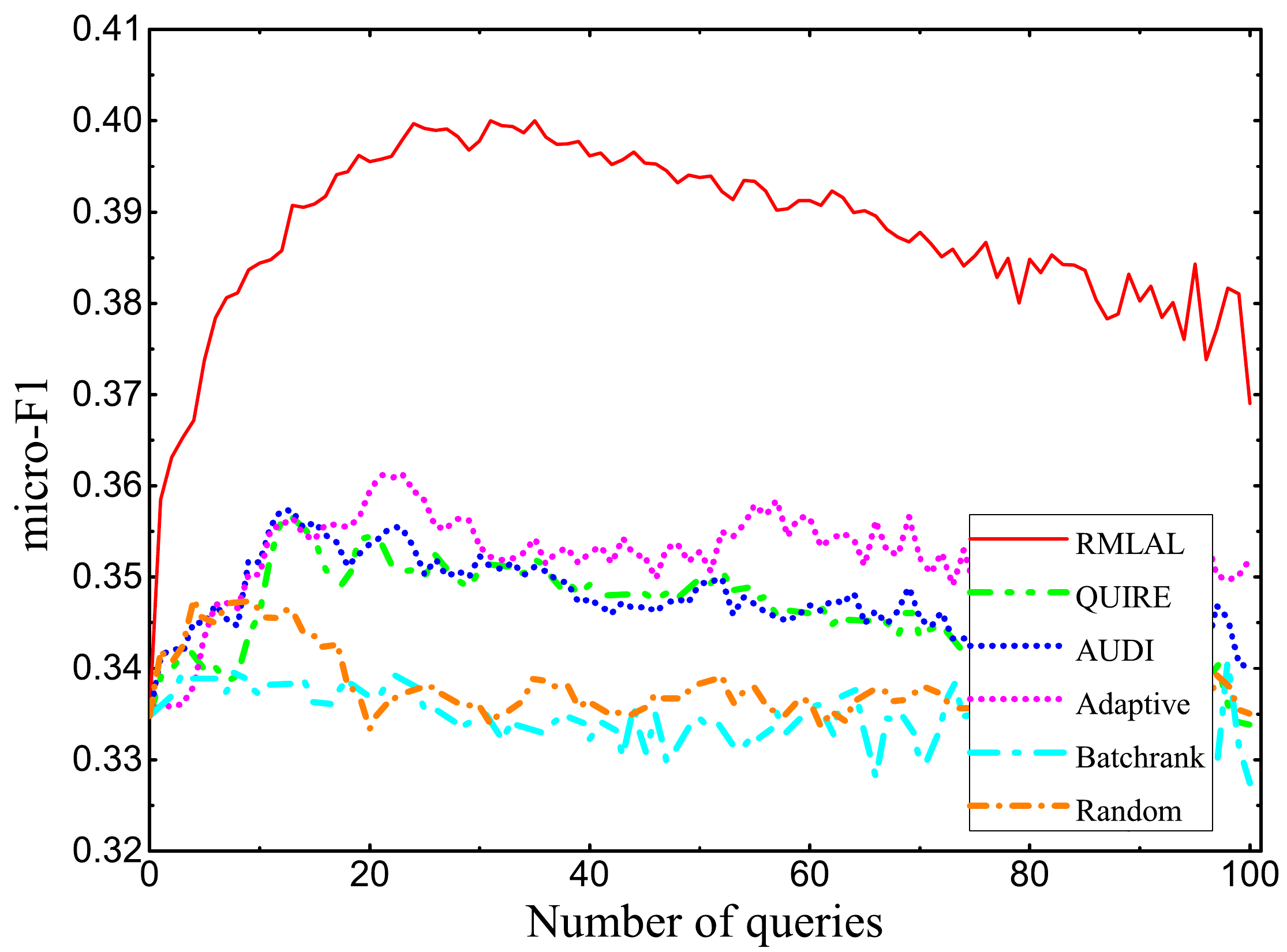, width=2.2 in,height=1.8 in}}
\caption{Comparison of different active learning methods on twelve benchmark datasets. The curves show the micro-F1 accuracy over queries, and each curve represents the average result of 5 runs.}
\end{center}
\end{figure*}
\subsection{Results}

We report the average results on each data set in Fig.4. Besides, we compare the competitors in each run with the proposed method based on the paired t-test at 95\% significance level, and show the Win/Tie/Lose for all datasets in Table II. From all these results, we can observe that the proposed method performs best on most of the data sets. It achieves the best results in almost the whole active learning process. In general, QUIRE and AUDI are two methods to query the label-instance pairs for labeling. They almost show the superior performance to the Batchrank and Adaptive, which query all labels for the instance. This demonstrates that querying the relevant labels is more efficient than querying all labels for one instance. But our method achieves the best performance by querying all the labels for one instance than querying the relevant label-instance pairs. The reason may be that although the Batchrank and Adaptive query all the labels, they cannot avoid the influence of the outlier labels without considering the labels correlation, leading to the query samples undesirable. This reason can also explain why Batchrank and Adaptive perform worse than random method on some data. For QUIRE and AUDI methods, some labels information is lost when they just query the limited relevant labels, and they need more samples to achieve a better performance. The results demonstrate the proposed method can not only achieve discriminative labeling but also avoid the influence of the outlier labels. To put it in nutshell, the proposed method merging the uncertainty and representativeness with MCC can solve the problems in multi-label active learning effectively as stated above.

For the computational cost, the time complexity of the proposed method is $O({t_1}{t_2}C(l+1)^3 + {t_2}u^2)$, where $t_1$ and $t_2$ are the number of iterations. The time complexity of Adaptive and AUDI are $O(Ctl^2)$ and $O((Cl)^2)$, respectively. QUIRE and Batchrank are costing, and the time complexity of them are both $O(u^3)$. $C$ is the number of classes. $u$ is the number of the unlabeled data. $l$ is the number of the labeled data and $t$ is the dimension of the data. Hence, compared with Adaptive and AUDI, the proposed method is costing, but it is relatively efficient when compared it with QUIRE and Batchrank. We show the time complexity of all the methods in Table III.
\begin{table}[htb]\label{table3}
\centering
\caption{THE TIME COMPLEXITY OF ALL THE METHODS}
\begin{tabular}{|c|c|} \hline
Methods&Time complexity\\ \hline
RMLAL& $O({t_1}{t_2}C(l+1)^3 + {t_2}u^2)$ \\ \hline
Adaptive& $O(Ctl^2)$ \\ \hline
AUDI &$O((Cl)^2)$ \\ \hline
Batchrank&$O(u^3)$ \\ \hline
QUIRE& $O(u^3)$ \\ \hline
\end{tabular}
\end{table}

\subsection{Evaluation parameters}
In the proposed method, the kernel parameter $\sigma$ is very important for the MCC, which controls all the robust properties of correntropy\cite{s23}. There are two tradeoff parameters on the uncertain part and representative part respectively. For convenience, in our experiments, we defined kernel size $\gamma= 1/(2*\sigma^2)$. Meanwhile, we fixed the kernel size as $1/C$ in label space , and fixed it as $1/t$ in feature space, where $t$ is the dimension of the data. To discover the influence of the kernel size for the proposed method, we evaluated the kernel size for MCC in label space. We reported the average results when the kernel size was set as $\{\gamma, 2\gamma, 4\gamma\}$ respectively on two popular benchmark datasets emotions and scene\cite{s20}, which had the same number of labels but with different LC. For the tradeoff parameters, we set them as $\beta_1 = \beta_2 = 1$. The other settings were same to the previous experiments.

\begin{figure}\label{fig5}
\begin{center}
\subfigure[Scene]{
\epsfig{file = 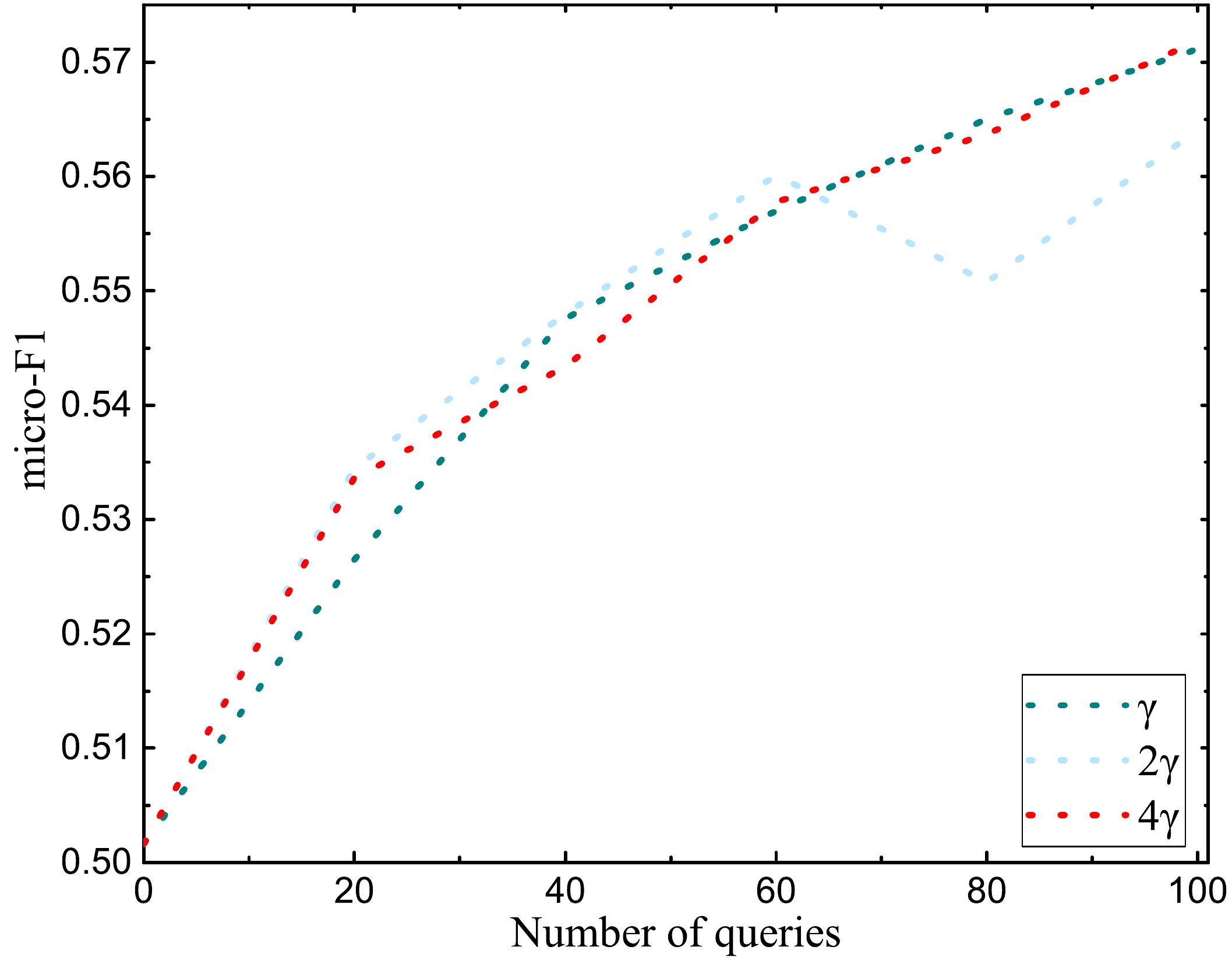, width=1.6 in,height=1.5 in}}
\subfigure[Emotions]{
\epsfig{file = 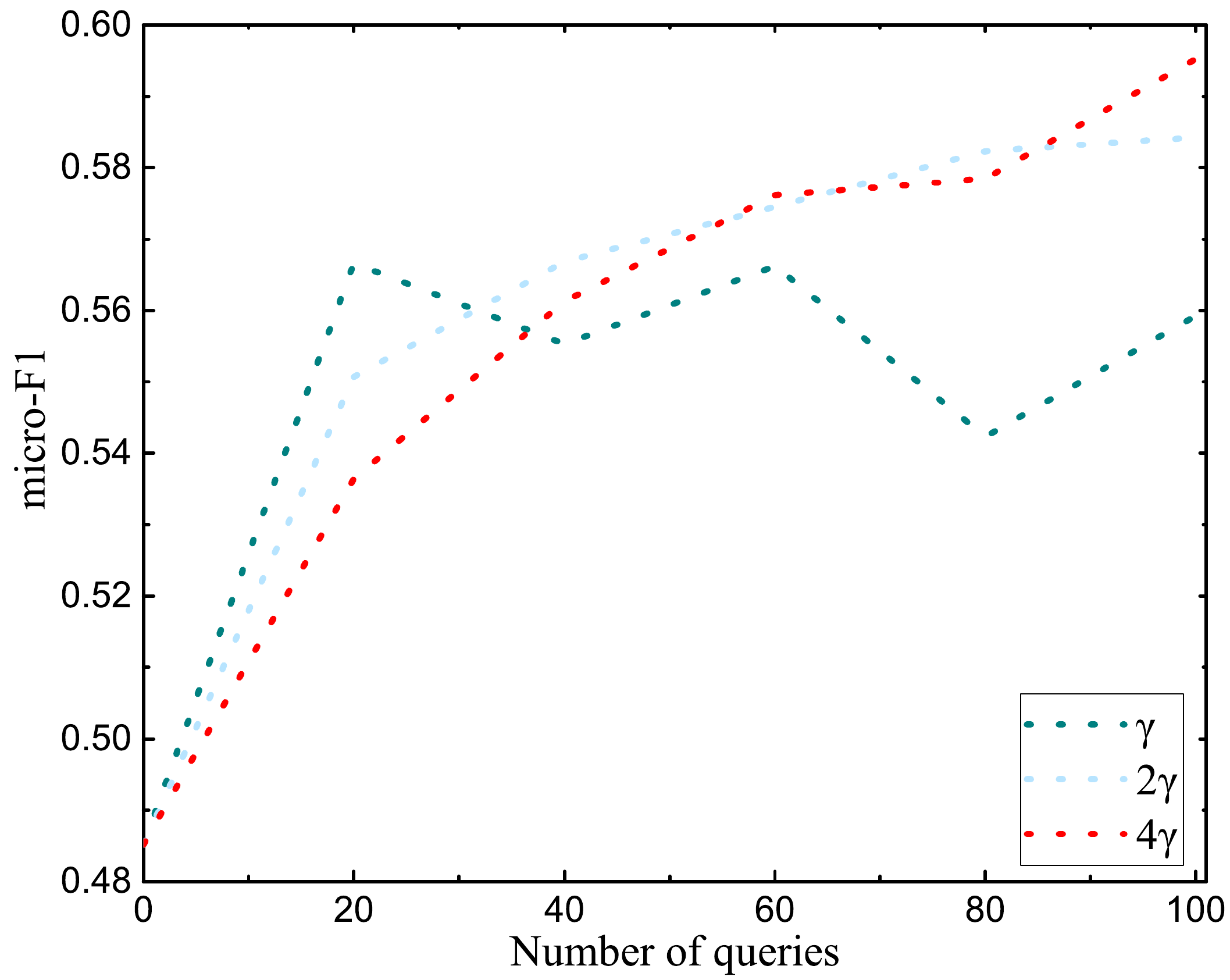, width=1.6 in,height=1.5 in}}
\caption{Comparison of different $\gamma$ on two data sets}
\end{center}
\end{figure}

\begin{figure}\label{fig6}
\begin{center}
\centering
\subfigure[Scene]{
\epsfig{file = 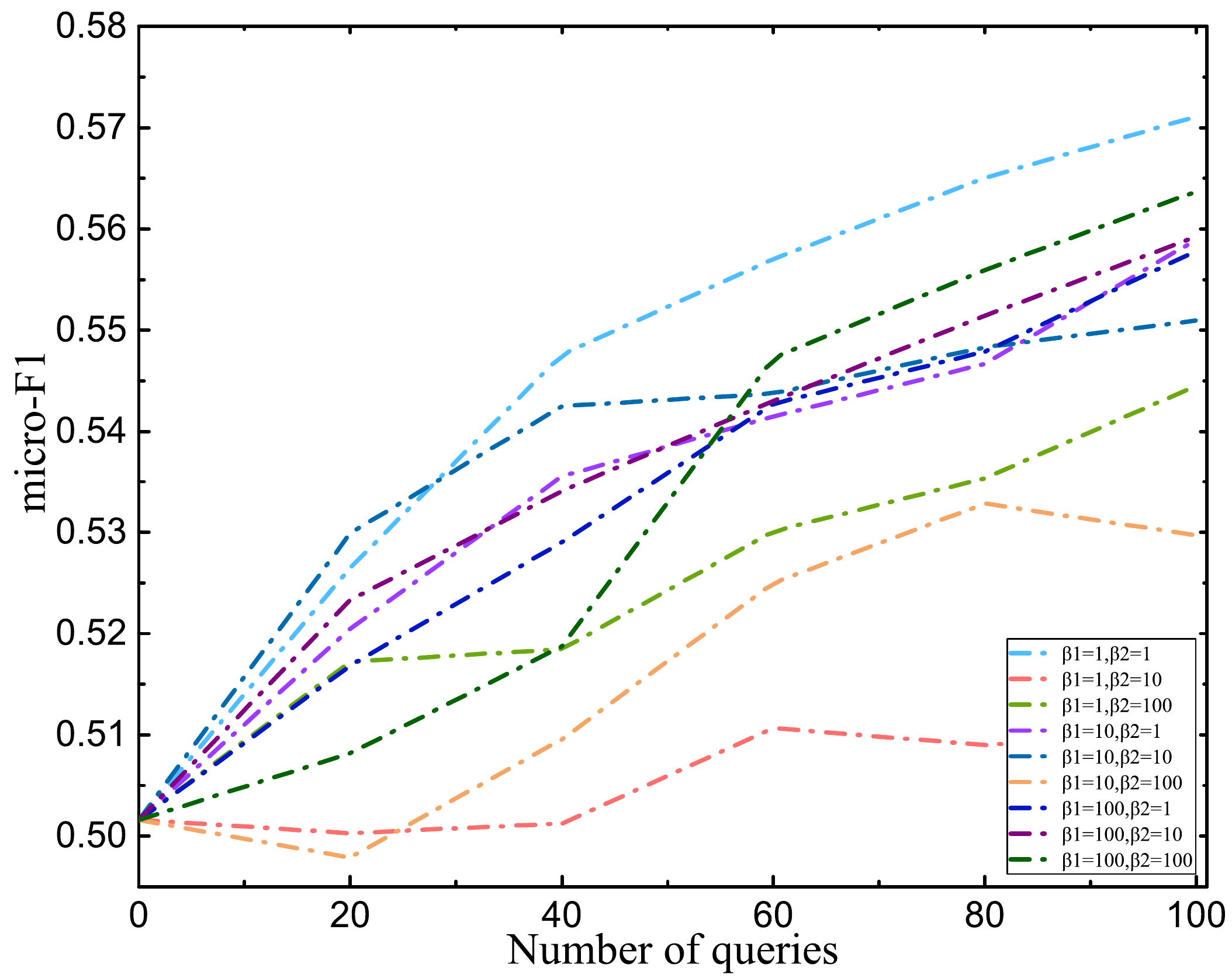, width=1.6 in,height=1.5 in}}
\subfigure[Emotions]{
\epsfig{file = 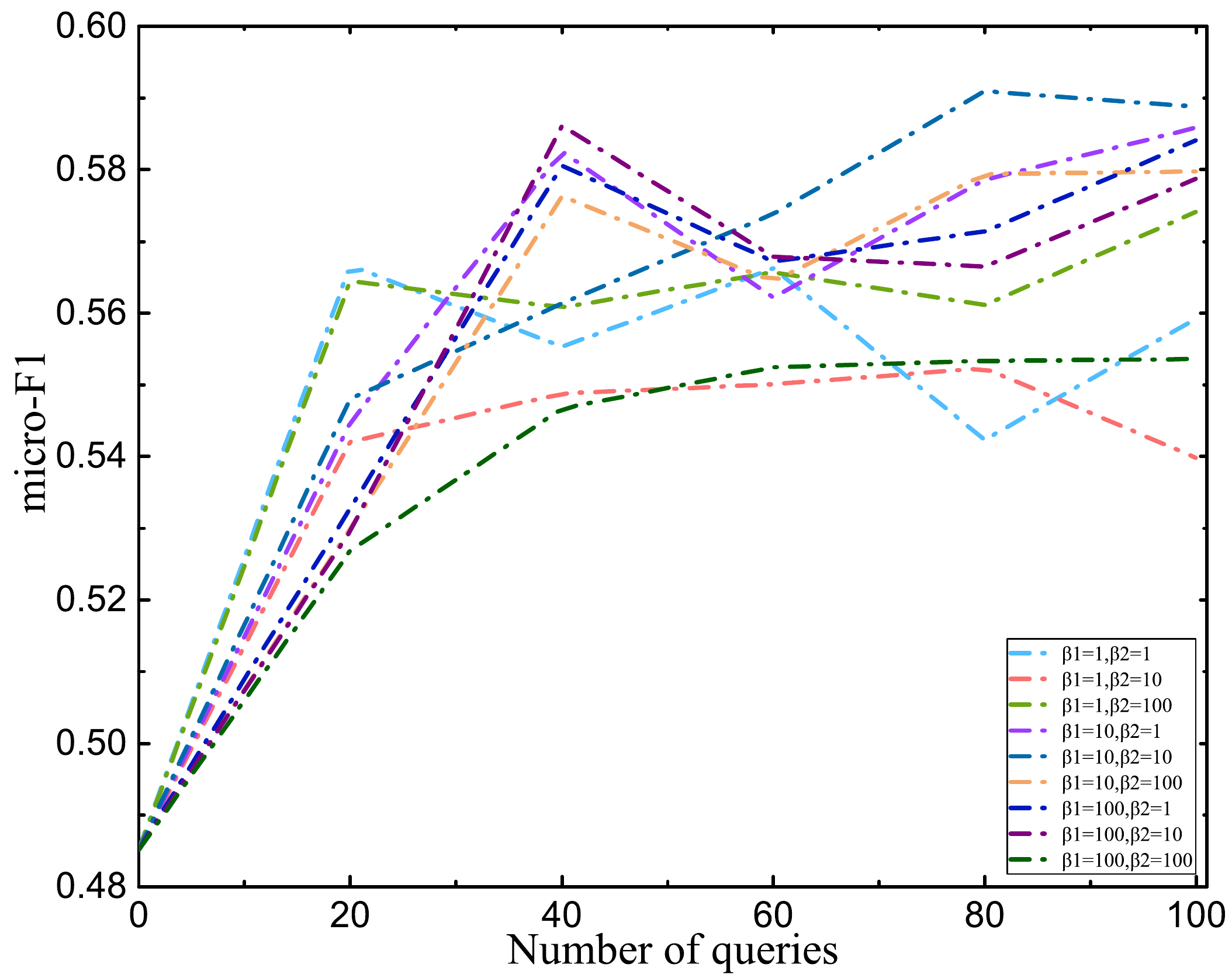, width=1.6 in,height=1.5 in}}
\caption{Comparison of different tradeoff parameter pairs$(\beta_1,\beta_2)$ on two data sets}
\end{center}
\end{figure}
  Fig. 5 shows the average results in 10 runs with the kernel size changing. From Fig. 5, we can observe that the larger the kernel size $\gamma$, the better results that the proposed method obtains. This may be because that when the kernel size is large, the values of the outlier labels based on MCC will be small in the objective function and the influence of the outlier labels is decreased as much as possible. Hence, we can set the kernel size $\gamma$ with a larger value for better performance. Fig.6 shows the average results in 10 runs with different pairs of the tradeoff parameters. For these results, we can observe that uncertain information and representative information have a big influence on the results. This may be because that the number of the labeled samples and that of unlabeled samples are changing in the active learning process. In active learning, the uncertain information is related to the labeled data, and the representative information is related to the unlabeled data. Hence, when the uncertain information and representative information are fixed, it is hard to control the required information in different iterations. From Fig.6, we can also observe that when $\beta_1$ is large and $\beta_2$ is small, the results on the two data sets are consistent, and they all achieve the relatively good results. Although the results are not the best, the proposed method performs stably and presents superiority to the results when $\beta_1$ is small and $\beta_2$ is large. Hence, in practice, a large value for $\beta_1$ and a small value for $\beta_2$ can be adopted so that the unlabeled data can be fully used.
  \subsection{Further Analysis}
 In order to further explain the motivation of the proposed method, we replaced the MCC loss function with MSE, which is usually adopted in the state-of-the-art methods\cite{s3,s21}. Meanwhile, a visual data set PASCAL VOC2007\footnote{http://host.robots.ox.ac.uk/pascal/VOC/voc2007/examples/index.html} was adopted. We selected a subset with 4666 samples and 20 classes from PASCAL VOC2007. The PHOW features and spatial histograms of each image were obtained with VLFeat toolbox\footnote{http://www.vlfeat.org/}. To observe the motivation directly, we show the images that are queried at the first, the twentieth, the fortieth, the sixtieth, the eightieth, and the one hundredth iterations in Fig. 7 and Fig. 8. Fig. 7 and Fig. 8 are obtained by the proposed method based on MCC and MSE respectively. The results obtained by MCC and MSE are also shown in the whole active learning process in Fig. 9. From Fig. 7, we can observe that the labels in each image are all very relevant to the image, and there are no outlier labels. Meanwhile, compared with the background, the object corresponding to the image's label covers a larger region in the image, leading to the result that the object is very relevant to the image. In Fig. 8, the leaves which are the background in the image selected at $20^{th}$ iteration cover a larger region than the bird which is a label to the image, and the mountain which is background in the image selected at $60^{th}$ iteration cover a larger region than the cow which is a label to the image. Hence, the labels of the images selected at the $20^{th}$ and $60^{th}$ iterations are less relevant to the images than the background is. Compared with the background, the labels of these images seem to be outlier labels. In a word, the method based on MSE may select the images that the backgrounds cover larger regions than the objects corresponding to the images' labels, while the proposed method based on MCC can decrease the influence of the outlier labels, and select the images that the labels are more obvious than the backgrounds to make full use of the labels in the images.

\begin{figure}\label{fig7}
\begin{center}
\centering
\subfigure[cat, TV monitor]{
\includegraphics[height=0.11\textwidth]{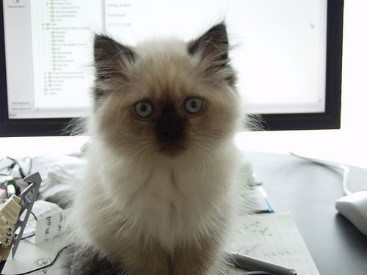}}
\subfigure[dining table, chair, bottle]{
\includegraphics[height=0.11\textwidth]{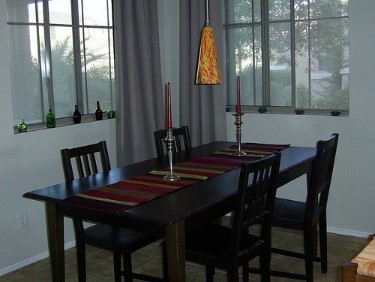}}
\subfigure[horse, person]{
\includegraphics[height=0.11\textwidth]{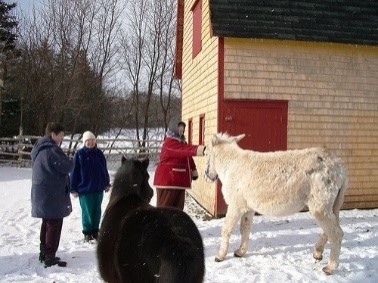}}
\subfigure[motorbike]{
\includegraphics[height=0.15\textwidth]{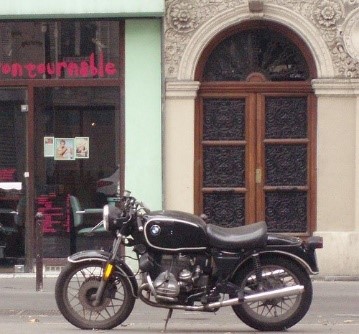}}
\subfigure[bicycle, person]{
\includegraphics[height=0.15\textwidth]{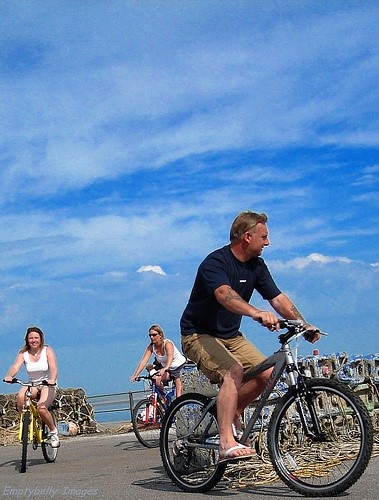}}
\subfigure[boat]{
\includegraphics[height=0.15\textwidth]{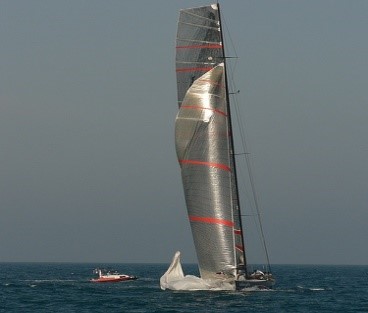}}
\caption{The images that are queried based on MCC at several iterations:(a) $1^{st}$ iteration; (b) $20^{th}$ iteration; (c) $40^{th}$ iteration;
(d) $60^{th}$ iteration; (e) $80^{th}$ iteration; (f) $100^{th}$ iteration}
\end{center}
\end{figure}

\begin{figure}\label{fig8}
\begin{center}
\centering
\subfigure[aeroplane]{
\includegraphics[height=0.1\textwidth]{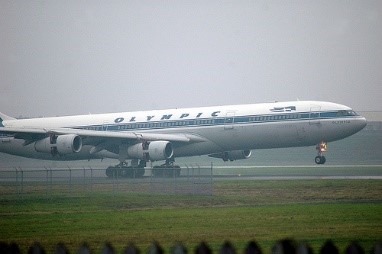}}
\subfigure[bird]{
\includegraphics[height=0.1\textwidth]{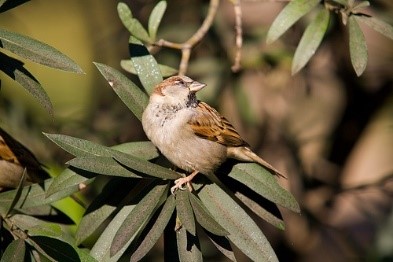}}
\subfigure[bird, person]{
\includegraphics[height=0.1\textwidth]{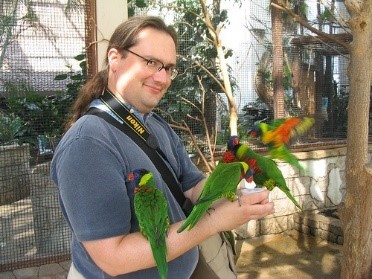}}
\subfigure[cow]{
\includegraphics[height=0.15\textwidth]{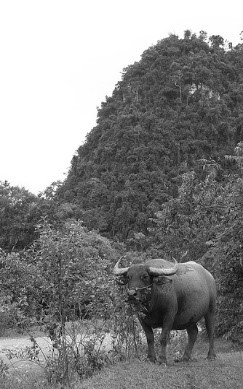}}
\subfigure[chair, dining table]{
\includegraphics[height=0.15\textwidth]{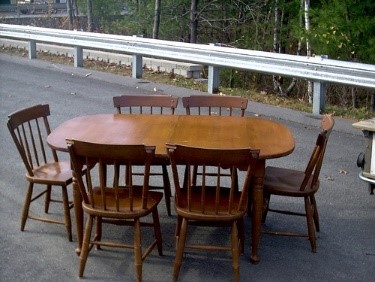}}
\subfigure[cat]{
\includegraphics[height=0.15\textwidth]{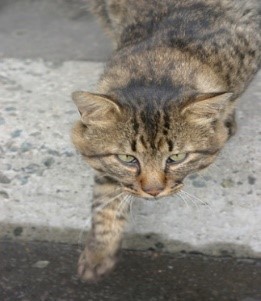}}
\caption{The images that are queried based on MSE at several iterations:(a) $1^{st}$ iteration; (b) $20^{th}$ iteration; (c) $40^{th}$ iteration;
(d) $60^{th}$ iteration; (e) $80^{th}$ iteration; (f) $100^{th}$ iteration}
\end{center}
\end{figure}
\begin{figure}\label{fig9}
\begin{center}
\epsfig{file = 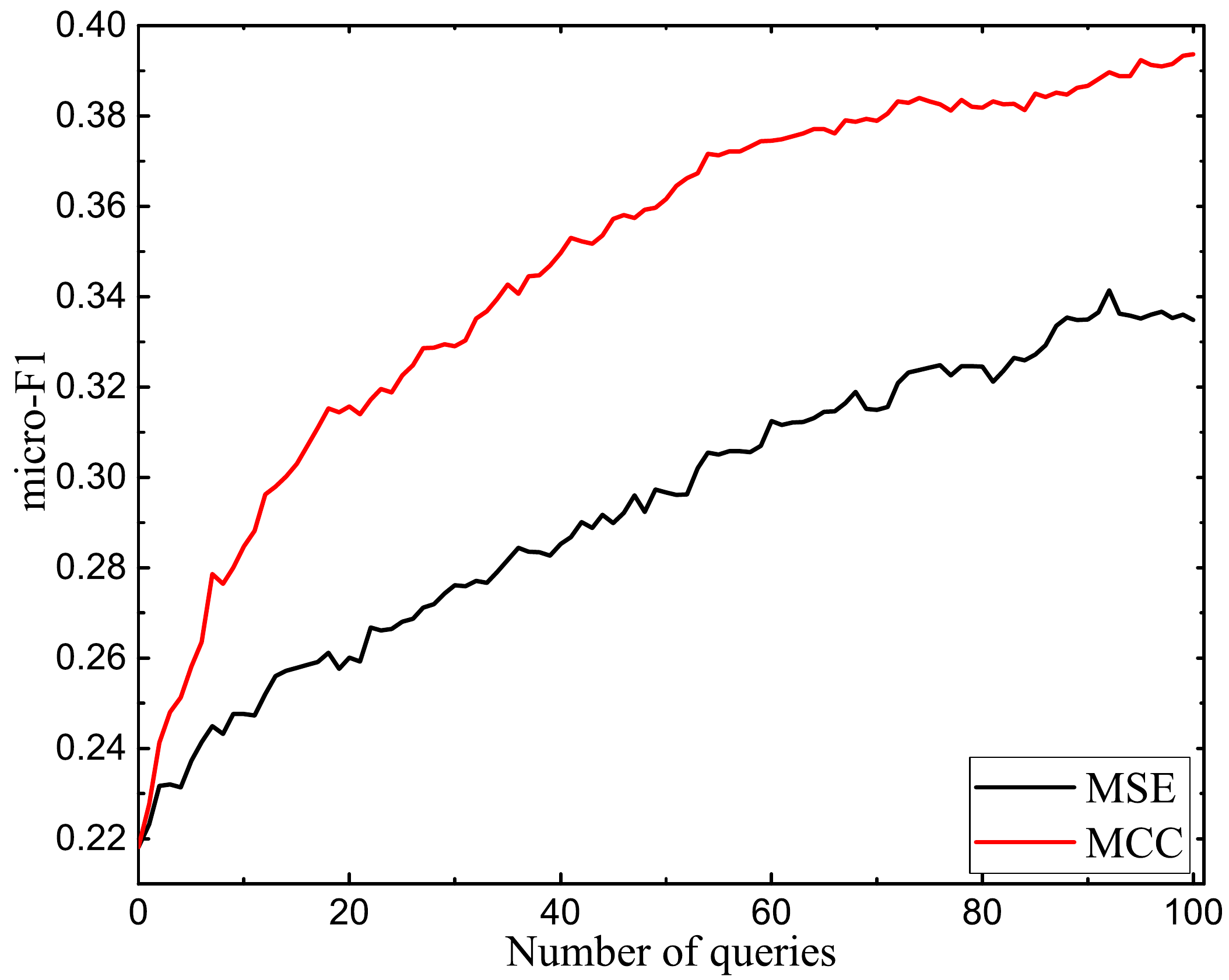, height=2.5 in}
\caption{The average results of the proposed method based on MCC and MSE in the whole active learning process}
\end{center}
\end{figure}

\section{Conclusion}
Outlier labels are very common in multi-label scenarios and may cause the supervised information bias. In this paper, we propose a robust multi-label active learning based on MCC to solve the problem. The proposed method queries the samples that can not only build a strong query model to measure the uncertainty but also represent the similarity well for multi-label data. Different from the traditional active learning methods by combining uncertainty and representativeness heuristically, we merge the representativeness into uncertainty with the prediction labels of the unlabeled data with MCC to enhance the uncertain information. With MCC, the supervised information of outlier labels will be suppressed, and that of discriminative labels will be enhanced. It outperforms state-of-the-art methods in most of the experiments. The experimental analysis also reveals that it is beneficial to update the tradeoff parameters that balance the uncertain and representative information during the query process. In our future work, we plan to develop an adaptive mechanism to tune these parameters automatically, making our algorithm more practical.


%

\ifCLASSOPTIONcaptionsoff
  \newpage
\fi



%

\bibliographystyle{IEEEtran}
\footnotesize
\bibliography{ref}

\end{document}